\pdfoutput=1  

\documentclass[10pt,twocolumn,letterpaper]{article}

\usepackage[pagenumbers]{cvpr} 

\usepackage{amsmath,amsfonts,bm}

\def\eqref#1{equation~\ref{#1}}

\def\1{\bm{1}}

\DeclareMathAlphabet{\mathsfit}{\encodingdefault}{\sfdefault}{m}{sl}
\SetMathAlphabet{\mathsfit}{bold}{\encodingdefault}{\sfdefault}{bx}{n}

\usepackage{flushend}

\definecolor{cvprblue}{rgb}{0.21,0.49,0.74}
\usepackage[pagebackref,breaklinks,colorlinks,allcolors=cvprblue]{hyperref}

\def\confName{CVPR}
\def\confYear{2027}

\DeclareRobustCommand{\Fluent}{\textup{\textsc{Fluent }}}

\DeclareRobustCommand{\revision}[1]{{\color{black}#1}}

\title{Vocabulary Aligned Explanations for Computer Vision Models}

\author{
Erfan Mirhaji \qquad Nikhil Krishnaswamy \qquad Jill Zarestky\\
Lisa Mason \qquad Nathaniel Blanchard \qquad Sarath Sreedharan
}

\begin{document}
\maketitle

\begin{abstract}
\label{abstract}
Optimizing vision models for accuracy does not force the model to learn and utilize human-understandable and domain-specific concepts and rules. We propose explaining image classifications through domain-specific concepts, visual evidence, and expert knowledge expressed as logical rules. We introduce \textsc{Fluent} (Framework for Localized, Uncertainty-Aware Explanations in Normative Terminology), a post-hoc framework that ranks expert-authored rules combining jointly required, alternative, and absent concepts. To account for prediction errors, \textsc{Fluent} combines predicted labels and concept states with empirically estimated likelihoods and concept priors. It scores each rule by calculating support over satisfying assignments to the concepts mentioned in that rule, then normalizes these scores to express relative confidence among candidate explanations. The selected rule is illustrated through occlusion-based highlights for positive concept mentions in the query image and localized examples in retrieved reference images for negative mentions. These illustrations are intended to show end users unfamiliar with the domain what the concepts named in an explanation look like. \textsc{Fluent} combines domain knowledge and model accuracy, and delivers visual concept highlights and explicit rule confidence to explain an image in the vocabulary of an application domain.
\end{abstract}

\section{Introduction}
\label{introduction}

Deep vision models can achieve high classification accuracy in visual tasks~\citep{resnet,ConvNeXtV2}, but label scores do not explain why a specific class was chosen. Optimizing classification accuracy does not require a model to express its predictions through concepts familiar to its users. 
This has led to a proliferation of highly capable but effectively black-box vision models.

Explainable artificial intelligence (XAI) methods were developed to explicitly address this shortcoming. Methods such as Grad-CAM or Integrated Gradients identify influential image regions~\citep{gradcam,integratedGradients}, while LIME fits local, interpretable approximations of prediction behavior~\citep{lime}. These methods can be used by AI experts to inspect where a model looks, but a highlighted region can contain several attributes; highlights alone do not specify what matters, whether the presence or absence of a feature supports the class label, or how different features interact with each other. As such, the usefulness of these methods for end-users have been under debate ~\citep{tcav}.

\begin{figure*}[t]
  \centering
  \includegraphics[width=0.72\textwidth]{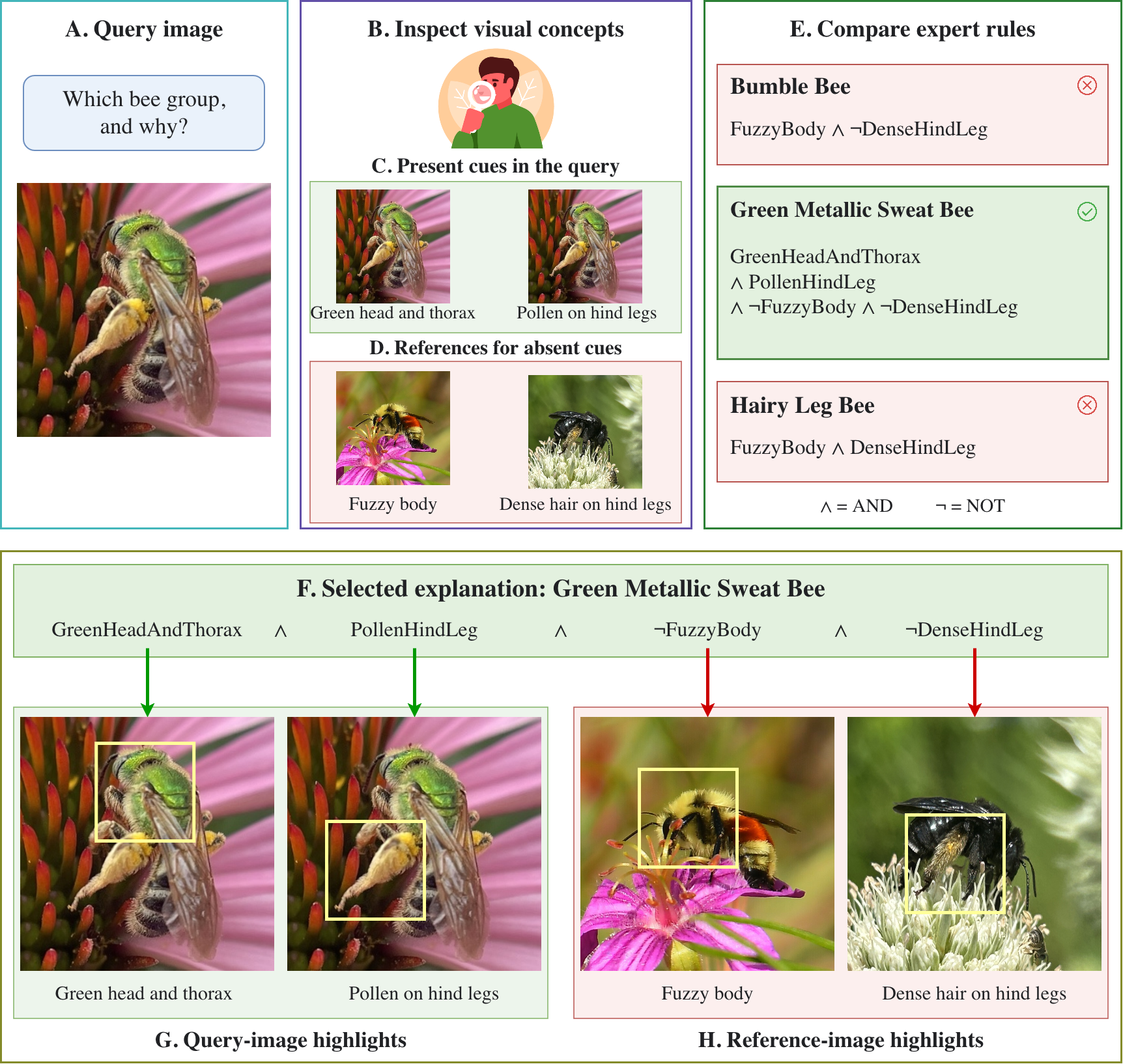}
    \caption{How a human expert would explain a bee image to someone in the domain who may be unfamiliar with its concepts and their appearance. \textbf{(A--C)}~The expert introduces the concepts and points out green coloration and pollen on the hind legs in the query. \textbf{(D)}~To explain fuzzy body hair and dense hind-leg hair, which are absent here, the expert shows reference images containing them. \textbf{(E--F)}~The expert connects these observations through the Green Metallic Sweat Bee rule. Here $\wedge$ means ``and'' and $\neg$ means ``not.'' \textbf{(G--H)}~Highlights guide the recipient to the relevant concepts in the query and references. This human explanation process is the running example for \textsc{Fluent}; Section~\ref{methodology} defines its automated procedure.}
  \label{flow}
\end{figure*}

To illustrate how these methods differ from human-generated explanations, consider an entomologist trying to explain a bee's identity to a non-expert (Figure~\ref{flow} provides a visual overview).
They set out with a set of domain-specific vocabulary or concepts, denoting relevant attributes (body hair, color, stripes, etc.).
The identity of the bee itself is determined by rules defined in terms of the absence or presence of these individual concepts.
For example, the picture in Figure~\ref{flow} corresponds to a green metallic bee because it has a green head and thorax, has pollen in its hind leg, doesn't have a fuzzy body, and doesn't have dense hair in its hind legs.
As part of the explanation, the expert would present the rule to the non-expert, along with examples that would teach them the individual concepts.
For positive concepts (\eg green head and thorax), the expert could rely on the current image, and for negative ones they might have to rely on images from alternative species that do exhibit that attribute.

While there exists some works that can support some steps of the explanatory workflow described above, no single work can completely replicate it.
Prominently, concept-based XAI methods introduce an interpretable vocabulary: TCAV measures concept sensitivity~\citep{tcav}, and Concept Bottleneck Models (CBMs) predict through explicit concepts~\citep{cbm}. Uncertainty-aware CBMs already model uncertainty in concept predictions~\citep{probcbm}, and structured prototype methods can reason with evidence for both presence and absence~\citep{prototree,protoargnet}. 
However, the primary missing piece is a mechanism that can support explanation that leverages complex logical rules over concepts.
Such a method should, for a given input, identify the most likely rule that can explain it.
Present the rule to the user and then help the user understand the individual concepts used within the rule.


We introduce \textsc{Fluent}, a post-hoc framework for explaining images in this manner using learned label and concept predictions. \textsc{Fluent} takes an expert-defined concept set and class-associated propositional rules as inputs. It trains concept predictors on labeled examples and estimates the error behavior of the label and concept predictors from labeled evaluation data. Independent estimation data are desirable; the bee splits do not fully meet this condition. The rule scorer operates on label and concept predictions and their estimated error rates, without requiring a particular backbone architecture or feature dimension. Our experiments use pretrained backbones with separately trained label and concept heads. Learned backbone parameters were intended to remain fixed, but inspected BatchNorm buffers differ from the pretrained checkpoints; Appendix~\ref{app:backbone_state} distinguishes fixed parameters from fixed inference state.

For a query image, \textsc{Fluent} combines the classifier’s predicted label with predictions about which domain concepts are present or absent to identify the best-supported domain rule. It accounts for the measured error rates of the label and concept predictors, weighing conflicting evidence according to their reliabilities. The selected rule is accompanied by a confidence score expressing its support relative to other candidate rules. The rule provides an interpretation of the predicted evidence using domain criteria; it is not a reconstruction of the classifier's internal decision process. We evaluate the accuracy of the classifier's label and the rule-implied label separately.

Following the expert's demonstration in Figure~\ref{flow}, \textsc{Fluent} pairs the selected rule with visual examples to show users what its concepts look like. For each positively mentioned concept, it highlights the query patch whose occlusion yields the lowest concept-predictor score. For negative concepts, it retrieves the most similar reference image containing that concept and highlights it there. In the bee example, query highlights show green coloration and pollen, while references show fuzzy body and dense hind-leg hair. These reference images illustrate the concepts named via negative literals.

Succinctly, our contributions are: \text{(I)} A post-hoc framework that explains images through domain-specific concepts and class-associated logical rules using features from a pretrained vision backbone. \text{(II)} A rule-scoring procedure that incorporates measured label and concept prediction errors and reports normalized support for competing explanations. \text{(III)} A visual grounding procedure that links positive concept mentions to query-image highlights and negative mentions to localized reference examples. \text{(IV)} An evaluation across 12 backbones and two benchmarks, examining rule recovery, rule-implied classification, concept-supervision budgets, and pretraining choices.  We evaluate \textsc{Fluent} across 12 pretrained vision backbones on a synthetic and a real  dataset (Appendix~\ref{app:results}).

\section{Related Work}
\label{related_work}
Saliency-based methods attribute predictions to input features by measuring pixel or region importance. Class saliency and class activation methods~\citep{simonyanDeepInsideCNN, gradcam,scorecam} highlight discriminative regions through class-specific gradients or forward-pass masking, while LIME~\citep{lime} uses perturbations to fit interpretable local approximations around individual predictions. Propagation and axiomatic methods, including Integrated Gradients~\citep{integratedGradients} and LRP~\citep{lrp}, attribute relevance using reference inputs or conservation rules. Attribution maps can be fragile: small, label-preserving perturbations can substantially alter saliency~\citep{fragileInterpretations}, and sanity checks find some methods insensitive to model parameters or training data~\citep{sanityChecks}. Connecting heatmaps to meaningful concepts requires human interpretation, limiting their usefulness for structured domain criteria and reasoning~\citep{surveyNNinterpretability}.

Concept-based methods bridge this gap with human-interpretable concepts. TCAV~\citep{tcav} measures class-score sensitivity to user-defined concept directions in activation space, while ACE~\citep{ace} reduces manual effort by discovering concepts through segmentation and clustering. Vision-language approaches associate internal units with natural-language descriptions, as in CLIP-Dissect~\citep{clipDissect}, or construct language-guided concept bottlenecks, as in LaBo~\citep{languageInBottle}. These methods provide concept vocabularies or scores; our explanation space instead consists of a supplied set of expert-authored logical rules.

Concept Bottleneck Models (CBMs)~\citep{cbm} route predictions through an explicit concept layer, enabling concept-level interventions for human correction at test time. Variants relax supervision or improve concept representations~\citep{posthocCBM,labelfree_cbm,cem}, while uncertainty-aware ProbCBMs~\citep{probcbm} propagate concept distributions into downstream decisions. CBMs typically learn concept-to-label mappings from data rather than encoding expert rules, and often require retraining or architectural constraints that may be undesirable for explaining off-the-shelf backbones.

Prototype networks explain decisions via similarity to prototypical parts or training instances. ProtoPNet~\citep{protopnet} learns class-specific prototype patches for localized explanations; variants improve compactness through prototype sharing and compositional reasoning~\citep{protopshare,prototree}. Recent work further targets human alignment and modern backbones~\citep{pipnet,protovit}. ProtoArgNet~\citep{protoargnet} reasons over prototypical evidence through structured argumentation, approaching rule-like explanations. ProtoTree and ProtoArgNet already support structured reasoning, including evidence against a prediction. 
We also rely on visual examples to ground concepts for users, but they aren't a complete explanation.

Contrastive and counterfactual methods ask how input changes would alter a decision~\citep{wachterCF,counterfactualVisual}, with diffusion-based generation~\citep{dvce} and semantically constrained steering~\citep{steex} improving realism. Pertinent negatives~\citep{pertinentNegatives} highlight features whose absence is decisive, a form of reasoning also relevant to expert rules. Negative evidence therefore has precedents; our method illustrates negatively mentioned domain concepts with reference examples.

We focus on combining uncertainty-aware selection among expert-defined logical rules with visual explanations of their concepts. The selected rule specifies how concept evidence is combined, while query-image highlights and reference examples communicate the rule's positively and negatively mentioned concepts, respectively.

\section{\textsc{Fluent} Method}
\label{methodology}
\textsc{Fluent} combines separately trained label and concept predictors with expert-authored rules. It ranks rules using empirical prediction-error likelihoods, normalizes their support, and illustrates positive and negative concept mentions through query and reference highlights.

\subsection{Labels, Concepts and Expert Rules}
\label{sec:labels_concepts_rules}

{We formalize the labels, concepts, and rules that the expert uses to explain the bee image to the recipient in Figure~\ref{flow}. Let $\mathcal L=\{l_1,l_2,\ldots,l_n\}$ be the class labels and $\mathcal X\subseteq\mathbb R^{h_{\mathrm{img}}\times w_{\mathrm{img}}\times3}$ the modeled domain of images, where $h_{\mathrm{img}}$ and $w_{\mathrm{img}}$ are the image height and width in pixels and the last dimension contains the three color channels. We train a domain-specific classifier head on features from a pretrained backbone, yielding the label classifier $\revision{\phi_{\mathrm{cls}}}:\mathcal X\to\mathcal L$. For an image $x\in\mathcal X$, its predicted label is $\hat\ell(x):=\revision{\phi_{\mathrm{cls}}}(x)$.}

{Domain experts specify the visual concepts used in the rules, such as green coloration of the head and thorax or pollen on the hind legs in Figure~\ref{flow}. Let $\mathcal C=\{c_1,c_2,\ldots,c_k\}$ be this concept vocabulary.} We treat each concept $c_i$ as a binary proposition about an image. For any image $x$, let $c_i(x)\in\{0,1\}$ denote the \emph{concept indicator} that concept $c_i$ is present in $x$: $c_i(x)\;:=\;\mathbf{1}\!\big[\,c_i \text{ is present in } x\,\big]\in\{0,1\}.$ We will use $\mathbf C(x)$ to denote the \emph{concept vector} for image $x$: $\mathbf C:\mathcal{X}\to\{0,1\}^k; \mathbf C(x) := \big(c_1(x),\ldots,c_k(x)\big).$ We use lowercase $\mathbf c\in\{0,1\}^k$ for a possible value of this vector.

We use propositional logic to combine present and absent concepts into expert rules. Absence of a concept matters, \eg under the supplied benchmark rule, a Bumble Bee must \emph{not} have dense hair on its hind legs. {Rules express concept presence and absence using literals.} {In propositional logic, a \emph{literal} is either a concept $c_i$ or its negation $\neg c_i$, meaning that the concept is absent.} Let $\mathrm{Lit}(\mathcal C)=\{c_1,\ldots,c_k,\neg c_1,\ldots,\neg c_k\}$ be the set of literals. \textsc{Fluent} assumes experts use \emph{propositional logical rules} to express why an image belongs to a class, in terms of which concepts are present or absent, \eg ``fuzzy hair on the body \emph{and} dense hair on the hind legs'' justifies Hairy Leg Bees. Such explanations combine several concepts, and might also include alternatives for a concept. We express each explanation as a formula over literals. The scoring definition accepts a rule $r$ in conjunctive normal form (CNF), that is, a conjunction of clauses, where each clause is a disjunction of literals: $
r=\bigwedge_{j=1}^{m}\big(\bigvee_{p=1}^{s_j}\ell_{j,p}\big),\quad \ell_{j,p}\in\mathrm{Lit}(\mathcal C).
$

Here $\land$ means AND and $\lor$ means OR: every clause must hold, and at least one literal must hold within each clause. For example, $(c_1\lor c_2)\land\neg c_3$ is CNF and requires either $c_1$ or $c_2$, together with the absence of $c_3$. We write $\mathbf c\models r$ when assigning concepts the values in $\mathbf c$ makes rule $r$ true. We write $\ell\sqsubseteq r$ when literal $\ell$ occurs in the written rule; this denotes occurrence, not logical implication.

{A class may have several rules describing different qualifying concept configurations.} For each class label $l\in\mathcal L$, experts provide a non-empty set of rules $\mathcal R_l$.
Let $\mathcal R_{expert}=\bigcup_{l\in\mathcal L}\mathcal R_l$ be the set of all expert-authored rules. {We assume that each supplied rule is assigned to only one class, so the rule sets are disjoint across classes:}
$\forall\, l_i\neq l_j\in\mathcal{L}:\; \mathcal R_{l_i}\cap \mathcal R_{l_j}=\varnothing.$

Based on this assumption, each rule belongs to exactly one class. We write $\ell(r)$ for the unique label such that $r\in\mathcal R_{\ell(r)}$. We also assume mutual exclusivity: no two distinct rules can be simultaneously satisfied by the same concept vector:
$
\forall\, r\neq r'\in \mathcal R_{expert},\ \forall\, \mathbf c\in\{0,1\}^k:\; {\mathbf{1}[\mathbf c\models r]+\mathbf{1}[\mathbf c\models r']\le 1}.$
{We also assume \emph{coverage}: every image in the domain satisfies a supplied rule. Together with mutual exclusivity, this means exactly one rule holds for each image:}
\begin{equation}
\label{eq_unique_rule}
\forall x\in\mathcal X:\quad \sum_{r\in\mathcal R_{expert}} \mathbf 1[\mathbf C(x)\models r] = 1.
\end{equation}

The evaluated rule sets use a narrower grammar: positive unit literals or positive disjunctions, together with unit negative literals. The explanation renderer is only supported for this grammar; it does not preserve arbitrary clauses mixing positive and negative literals. The general scoring definition does not imply a validated general-CNF renderer.

\subsection{The Explanation Problem}
\label{sec:explanation_problem}

{Given an image $x$, the explanation problem is to recover the unique supplied rule satisfied by its true concept vector $\mathbf C(x)$ (Equation~(\ref{eq_unique_rule})). Since the true concepts are unknown at inference time, \textsc{Fluent} scores candidate rules using predicted labels and concept states.}

\subsection{Concept Predictors}
\label{sec:concept_predictors}

For each concept $c_i$, we train a predictor $\revision{\phi_{c_i}}:\mathcal X\to[0,1]$ to infer its presence in a new image. Its score is used for localization, while its binary prediction $\hat c_i(x)$ is used for rule inference; we collect these predictions in $\hat{\mathbf C}(x)=(\hat c_1(x),\ldots,\hat c_k(x))$. Writing $\Pr$ for probability, we estimate each predictor's conditional likelihoods $\Pr(\hat c_i\mid c_i)$ from its prediction counts on the designated concept-labeled test set (Table~\ref{table_conditional_probs}). For $h,t\in\{0,1\}$, $\Pr(\hat c_i=h\mid c_i=t)$ is the probability that concept predictor $i$ outputs $h$ when the true concept state is $t$. The estimation procedure and conditional-likelihood table are given in Appendix~\ref{app:concept_likelihoods}.

\subsection{Selecting the Best Rule via Probabilistic Modeling}
\label{sec:rule_selection}

\begin{figure}[!tb]
  \centering
  \begin{minipage}[c]{0.34\linewidth}
  \includegraphics[width=\linewidth,trim=0 27bp 0 30bp,clip]{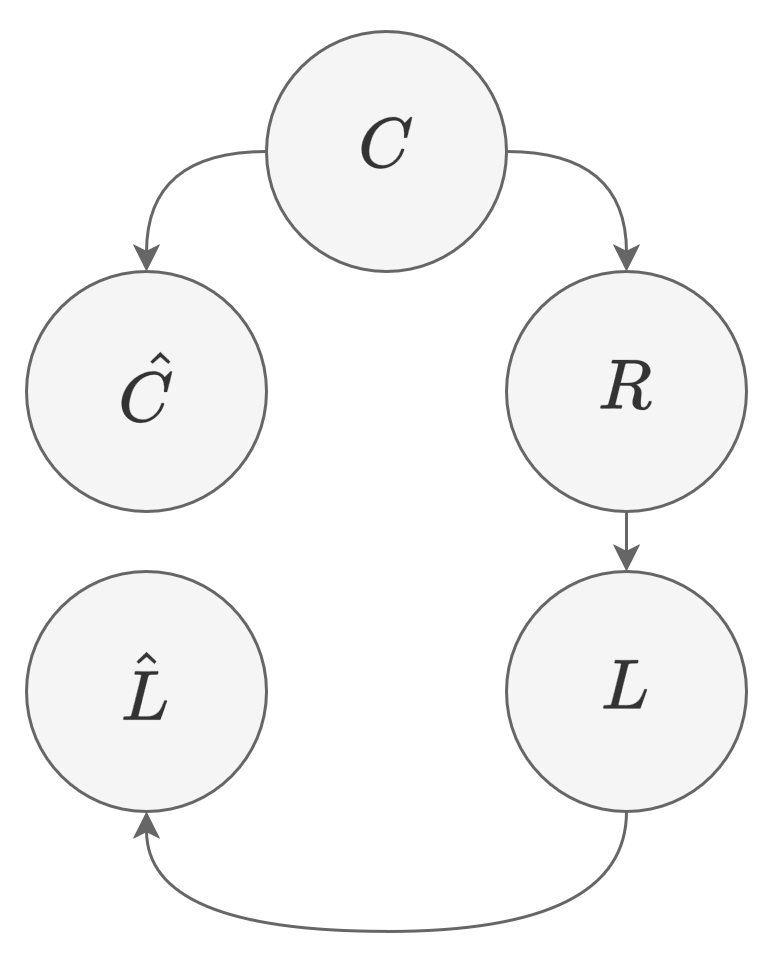}
  \end{minipage}\hfill
  \begin{minipage}[c]{0.62\linewidth}
  \caption{{Bayesian model for rule selection. True concepts $C$ determine a rule $R$ and its label $L$; the observed predictions $\hat C$ and $\hat L$ are modeled through concept and label prediction likelihoods. Implemented scores use only the concept predictions mentioned by each candidate rule.}}
  \label{fig_bayesnet}
  \end{minipage}
\end{figure}

{The label and concept predictors can make errors, so rule selection combines their observed outputs with empirical estimates of their reliability. Figure~\ref{fig_bayesnet} gives the probabilistic reference model: the true concept vector $C$ determines the satisfied rule $R$, which determines its class $L$. The predicted concepts $\hat C$ depend on $C$, and the predicted label $\hat L$ depends on $L$. These dependencies separate the logical rule from the prediction errors used to score it.}
At inference time, given an image $x$ we observe $\hat C=\hat{\mathbf C}(x)$ and $\hat L=\hat \ell(x)$. {Below, lower-case values denote assignments to these random variables, and we suppress $x$ when writing the observed predictions.}

{Let $\Omega=\{\mathbf c\in\{0,1\}^k:\exists r\in\mathcal R_{expert},\ \mathbf c\models r\}$ be the covered concept assignments, and let $A=\{C\in\Omega\}$ be the coverage event. Mutual exclusivity ensures that each assignment in $\Omega$ satisfies exactly one rule. We first approximate concept prevalence by an independent product $q(\mathbf c)$, with $\Pr(c_i=1)=\pi_i$ and $\Pr(c_i=0)=1-\pi_i$, where $\pi_i$ is estimated from concept metadata on a sample of the label-training data. The reference model uses this product conditioned on $A$; conditioning can make the concepts dependent. Its assumptions are:}

\begin{align}
q(\mathbf c)&=\prod_{i=1}^k\Pr(c_i)\label{eq:pr_c}\\[-3pt]
&\qquad\text{(I) independent base prior},\notag\\
\Pr(\hat{\mathbf C}\mid\mathbf c)&=\prod_{i=1}^k\Pr(\hat c_i\mid c_i)\label{Chat_given_c}\\[-3pt]
&\qquad\text{(II) separate concept channels},\notag\\
\Pr(\hat\ell\mid l,r,\mathbf c,\hat{\mathbf C})&=\Pr(\hat\ell\mid l)\label{eq:label_conf}\\[-3pt]
&\qquad\text{(III) label-only error channel},\notag\\
\Pr(r\mid\mathbf c,A)&=\mathbf1[\mathbf c\models r],\quad\mathbf c\in\Omega\label{eq_r_given_C}\\[-3pt]
&\qquad\text{(IV) deterministic rule},\notag\\
\Pr(l\mid r)&=\mathbf1[l=\ell(r)]\label{eq_l_given_r}\\[-3pt]
&\qquad\text{(V) deterministic label}.\notag
\end{align}

{Assumption (II) means each concept prediction depends only on its own true concept state; (III) means the label prediction depends only on the true class. The concept likelihoods come from concept-test prediction counts (Appendix~\ref{app:concept_likelihoods}), and the label likelihoods from the Label Test confusion matrix. These independence assumptions are modeling approximations, not consequences of the predictors sharing a backbone. Appendix~\ref{app:full_rule_posterior} derives the posterior that conditions on all concept predictions under this reference model.}

{The implemented rule score uses a candidate-specific subset of these predictions. Define the \emph{mentioned concept set} as
$\mathcal I_{\rho}(r):=\{i\in\{1,\ldots,k\}:c_i\sqsubseteq r\ \lor\ \neg c_i\sqsubseteq r\}$.
Rule satisfaction depends only on the true concepts with indices in $\mathcal I_{\rho}(r)$. For scoring, we additionally choose to use only their predictions $\hat{\mathbf C}_r=(\hat c_i)_{i\in\mathcal I_{\rho}(r)}$, omitting the prediction likelihoods for other concepts. This choice defines the score used in the experiments; it is not an algebraic simplification of the full-evidence posterior. The unnormalized score is}

\begin{multline}
\label{eq_rule_post_restricted}
w(r,x)=\Pr(\hat\ell\mid\ell(r))
\sum_{\mathbf c_{\mathcal I_{\rho}(r)}\in\{0,1\}^{|\mathcal I_{\rho}(r)|}}\\
\mathbf1[\mathbf c_{\mathcal I_{\rho}(r)}\models r]
\prod_{i\in\mathcal I_{\rho}(r)}\Pr(\hat c_i\mid c_i)\Pr(c_i).
\end{multline}

{The indicator equals one when the assignment satisfies every clause in $r$, and zero otherwise. Thus, the sum accounts for all satisfying assignments to the mentioned concepts, weighted by their prevalence and the likelihood of the observed predictions. The label factor accounts for how often a true instance of class $\ell(r)$ receives the observed label $\hat\ell$. We normalize these weights over the supplied rules and select the largest:}

\begin{equation}
\label{eq_r*_final}
\begin{gathered}
\operatorname{Conf}(r,x)=\frac{w(r,x)}{\sum_{r'\in\mathcal R_{expert}}w(r',x)},
\\
r^*(x)=\operatorname*{arg\,max}_{r\in\mathcal R_{expert}}\operatorname{Conf}(r,x).
\end{gathered}
\end{equation}

{When the denominator is positive, the confidences sum to one and express each rule's share of the total score. A two-concept rule and a four-concept rule use different subsets of the predictions, so these normalized scores are not posterior probabilities based on the same observed evidence. Appendix~\ref{app:full_rule_posterior} gives the exact difference from the full posterior and the implemented fallback when all weights are zero.}

\subsection{Highlighting Concepts}
\label{sec:highlighting}

{After selecting $r^*(x)$, \textsc{Fluent} illustrates its concepts as the expert does for the recipient in Figure~\ref{flow}(G--H), helping people unfamiliar with the concepts connect their names to visible features. Positive mentions receive query-image highlights; negative mentions receive highlights in reference images containing the concept. Define the positive and negative occurrence sets as $\mathcal I_{\rho}^+(r):=\{i:c_i\sqsubseteq r\}$ and $\mathcal I_{\rho}^-(r):=\{i:\neg c_i\sqsubseteq r\}$.} For $i\in\mathcal I_{\rho}^+(r^*(x))$, we highlight the query-image patch whose occlusion produces the lowest concept-predictor score. For $i\in\mathcal I_{\rho}^-(r^*(x))$, we retrieve a similar reference image containing $c_i$ and highlight that concept there. These displays illustrate the literals mentioned in the rule. In a clause such as $c_1\lor c_2$, only one alternative must hold, so highlighting both concepts does not assert that both are present.

\subsubsection{Highlighting Positive Concepts}
\label{Highlighting_Positive_Concepts}

We localize a positively mentioned concept by testing how masking different image patches changes its predictor's score. For clarity, consider a single $N\times N$ grid of non-overlapping patches in $x\in\mathcal X\subseteq\mathbb R^{h_{\mathrm{img}}\times w_{\mathrm{img}}\times3}$. Here $h_{\mathrm{img}}$ and $w_{\mathrm{img}}$ are the image height and width, distinct from the predictor notation $\revision{\phi_{c_i}}$. Choose $N$ that divides both dimensions, and set the patch height $p_h=h_{\mathrm{img}}/N$ and width $p_w=w_{\mathrm{img}}/N$. Using integer pixel coordinates $(u,v)$ starting at zero, patch $(a,b)$ contains the coordinates: $ P_{a,b}=\{(u,v):(a-1)p_h\le u<ap_h,\ (b-1)p_w\le v<bp_w\}, \quad a,b\in\{1,\ldots,N\}. $




Let $x_{u,v}\in\mathbb R^3$ denote the color vector at pixel $(u,v)$. To test patch $(a,b)$, we replace its pixels by black, producing
$x^{(a,b)}_{u,v}=x_{u,v}\,\mathbf{1}\!\left[(u,v)\notin P_{a,b}\right]$.
We evaluate $\revision{\phi_{c_i}}$ on each occluded image and select the patch attaining the lowest score. Keeping that patch and masking the rest defines the highlight $x_{c_i}^*$:
$
(a^*,b^*)=\operatorname*{arg\,min}_{a,b\in\{1,\ldots,N\}}\revision{\phi_{c_i}}(x^{(a,b)}), \;
(x_{c_i}^*)_{u,v}=x_{u,v}\,\mathbf{1}\!\left[(u,v)\in P_{a^*,b^*}\right].
$

\subsubsection{Highlighting Negative Concepts}
\label{Highlighting_Negative_Concepts}

{A reference image can illustrate a concept that the selected rule mentions negatively, as in Figure~\ref{flow}(H). For each such concept $c_i$, \textsc{Fluent} uses a reference dataset containing that concept:
$\mathcal D^{ref}_{c_i}=\{x_j\}_{j=1}^{M_{c_i}},\quad c_i(x_j)=1\ \forall j$.} For each $\neg c_i$ with $i \in \mathcal I_{\rho}^-(r^*(x))$, \textsc{Fluent} retrieves the most similar image $x^\sim_{c_i} \in \mathcal D^{ref}_{c_i}$ to image $x$ using cosine similarity in the backbone feature space: $x^\sim_{c_i} = \operatorname*{arg\,max}_{x_j \in \mathcal D^{ref}_{c_i}} \operatorname{similarity}(x, x_j)$. In \textsc{Fluent}, the similarity function is defined as: $\text{similarity}(x, x_j) = \frac{z \cdot z_j}{\|z\| \|z_j\|}$.
Here, $z=f_\lambda(x)$ and $z_j=f_\lambda(x_j)$ where $f_\lambda:\mathcal X\to\mathbb R^{d_\lambda}$ maps an image to its $d_\lambda$-dimensional feature vector at backbone layer $\lambda$ (typically the penultimate layer). After retrieving $x^\sim_{c_i}$, \textsc{Fluent} applies the same occlusion procedure to illustrate $c_i$ in that reference image. We denote the resulting highlight in the reference image by $x_{c_i}^{\sim *}$. We collect visual highlights into:
$\mathcal H^+(x) := \{x_{c_i}^* : i \in \mathcal I_{\rho}^+(r^*(x))\}, \; \mathcal H^-(x) := \{x_{c_i}^{\sim *} : i \in \mathcal I_{\rho}^-(r^*(x))\}.$ {These two sets correspond to the query and reference highlights illustrated in Figure~\ref{flow}(G--H).} This cosine comparison assumes nonzero feature norms. A faithful implementation must preserve the original query feature vector across all occlusion passes. The qualitative example illustrates the display and does not establish the correctness of the reference ranking (Appendix~\ref{app:explanation_example}).

\subsection{Explanation Tuple}
\label{FLUENT_Explanations}

\textsc{Fluent} combines the predicted label, selected rule, normalized rule score, and two sets of visual highlights into the explanation tuple: $
\mathcal{EXP}(x) = \Big(\hat{\ell}(x), r^*(x), \operatorname{Conf}(r^*(x),x), \mathcal{H}^+(x), \mathcal{H}^-(x) \Big)$. Here $\hat\ell(x)=\revision{\phi_{\mathrm{cls}}}(x)$ is the classifier label, $r^*(x)$ is the selected rule, $\operatorname{Conf}(r^*(x),x)$ is its normalized score, and $\mathcal H^+(x)$ and $\mathcal H^-(x)$ illustrate its positive and negative concept mentions, respectively. A complete synthetic explanation is shown in Figure~\ref{fig_FLUENT} in Appendix~\ref{app:explanation_example}.


\section{Experiments}
\label{sec:experiments}

We report experiments spanning 12 pretrained vision backbones with separately trained label heads and linear concept predictors. The evaluation measures agreement with supplied rules. The experiments use one configured seed and have the split-independence limitations described below.

\textbf{Benchmarks.}
The synthetic benchmark has three classes, 16 concepts, and six rules; generated concept metadata determine the unique target rule. The evaluation does not isolate concept evidence from possible background shortcuts. The bee benchmark has three groups, four concepts, and one rule per class. Its Explanation Test concept vectors are class-constant templates, without independent per-image verification of visible concepts. Thus bee RuleAcc measures agreement with those templates, not verified recovery of visible concepts. Rules and splits appear in Appendix~\ref{app:datasets}.

\textbf{Protocol.}
The Explanation Test sets contain 300 synthetic and 600 bee images. The \emph{synthetic reference configuration} uses $T{=}5000$ label-training images per class, $C{=}500$ positive and 500 negative training examples per concept, and prior-estimation sample fraction $P{=}0.1$. Thus $C$ counts examples per polarity per concept, and $P$ controls the fraction of label-training data sampled to estimate concept prevalence. The \emph{bee reference configuration} uses fixed dataset splits and $P{=}0.1$; synthetic $T$ and $C$ budgets do not apply to bees. Appendices~\ref{app:synthetic_results} and~\ref{app:real_results} give all configurations. Likelihood estimates use the Label Test and Concept Test confusion matrices. Appendix~\ref{app:protocol} details the experimental settings, inference-state checks, and reproducibility limitations.

\textbf{Metrics.}
LabelAcc measures classifier accuracy, RuleAcc measures agreement with the target rule, and RuleLabelAcc measures accuracy of the selected rule's implied label. Selecting the other rule of a correct synthetic class is an error only for RuleAcc; the metrics coincide on bees because each class has one rule. ConceptAcc averages binary accuracy on the Concept Test sets. Avg~$\operatorname{Conf}$ averages normalized rule support and does not measure calibration. Full definitions are in Appendix~\ref{sec:eval_metrics}.

\section{Results}
\label{sec:results}

Table~\ref{tab:headline_results} reports all five evaluation metrics for the same four representative backbones on both datasets, covering classic and modern convolutional models, plain and hierarchical transformers. Complete results for all 12 backbones are in Appendices~\ref{app:synthetic_results} and~\ref{app:real_results}. The main findings are that rule recovery depends on concept quality as well as class prediction, and that the effect of rule selection on label accuracy differs across datasets. Concept evidence improves rule-implied labels for the weak synthetic ResNet classifiers, whereas rule-implied labels are less accurate than classifier predictions for every backbone on bees.

\begin{table}[!ht]

\centering
\caption{Four representative backbones in the reference configurations: $T{=}5000$, $C{=}500$, $P{=}0.1$ for the synthetic benchmark; fixed dataset splits and $P{=}0.1$ for bees. Accuracies are percentages; Avg~$\operatorname{Conf}$ is unitless. ConceptAcc averages accuracy on the concept test sets; the other metrics use the Explanation Test splits (300 synthetic and 600 bee images). Bold marks column maxima among the displayed models within each dataset. Bee values have the overlap limitations in Appendix~\ref{app:evaluation_limits}. Full results appear in Appendix Tables~\ref{tab:main_results} and~\ref{tab:real_auto}.}
\label{tab:headline_results}
\setlength{\tabcolsep}{3pt}
\small
\begin{tabular}{lrrrrr}
\toprule
Backbone & Label & Concept & Rule & RuleLabel & Avg \\
         & Acc   & Acc     & Acc  & Acc       & $\operatorname{Conf}$ \\
\midrule
\multicolumn{6}{l}{\textit{Synthetic}} \\
RN50 (TV)       & 39.0 & 79.9 & 69.7 & 85.3 & 0.634 \\
CNXv2-L         & 99.3 & 93.3 & 93.7 & \textbf{99.7} & 0.925 \\
ViT-B/16 (DINO) & \textbf{99.7} & 92.8 & \textbf{96.7} & \textbf{99.7} & 0.935 \\
Swin-B (IN-22k) & 99.0 & \textbf{95.2} & 94.7 & 99.0 & \textbf{0.949} \\
\midrule
\multicolumn{6}{l}{\textit{Bees}} \\
RN50 (TV)       & 91.2 & 85.5 & 87.8 & 87.8 & 0.770 \\
CNXv2-L         & \textbf{99.7} & 96.4 & 85.8 & 85.8 & \textbf{0.967} \\
ViT-B/16 (DINO) & 97.7 & 96.2 & 84.3 & 84.3 & 0.927 \\
Swin-B (IN-22k) & \textbf{99.7} & \textbf{97.7} & \textbf{98.2} & \textbf{98.2} & 0.947 \\
\bottomrule
\end{tabular}
\end{table}

\subsection{Synthetic Benchmark}
\label{sec:synthetic_results}
\label{sec:synthetic_findings}

\textbf{\revision{Rule recovery varies across backbones.}}
\revision{Under the synthetic reference configuration, ViT-B/16 with DINO achieves the highest rule accuracy, $96.7\%$. Accurate class predictions alone are insufficient: ConvNeXt~V2-L with supervised fine-tuning reaches $93.7\%$ rule accuracy, compared with $54.0\%$ for its FCMAE-only variant, although both exceed $94\%$ label accuracy. Their mean concept accuracies differ substantially ($93.3\%$ versus $56.9\%$). Comparisons of pretraining variants are descriptive, not controlled causal estimates. Complete configuration results are in Appendix~\ref{app:synthetic_results}.}

\textbf{Rule-implied labels can correct classifier errors.}
\revision{For ResNet-50 (TV), combining concept and label evidence raises rule-implied label accuracy to $85.3\%$ from the classifier's $39.0\%$. Across the three ResNet backbones, gains are $26.7$--$46.3$ percentage points under the synthetic reference configuration. These gains concern the label inferred from the selected rule; \Fluent retains the original classifier prediction. For the nine backbones with at least $90\%$ classifier accuracy, the two label accuracies differ by at most $0.7$ points under the synthetic reference configuration. Appendix~\ref{app:full_sweep} provides every configuration.}

\subsection{Bee Benchmark}
\label{sec:real_world}
\label{sec:real_auto_results}

\textbf{Strong rule recovery does not guarantee label correction.}
\revision{Under the bee reference configuration, rule agreement with the class-derived templates on field photographs ranges from $43.8\%$ to $98.2\%$, with Swin-B (IN-22k) performing best. Unlike the synthetic setting, rule-implied label accuracy is below the original classifier's accuracy for every backbone, by $1.5$--$21.8$ percentage points. For ConvNeXt~V2-L, the gap is $99.7\%$ label accuracy versus $85.8\%$ rule accuracy. Thus, concept evidence can also introduce errors. Concept accuracy measured on curated exemplars need not transfer to field photographs, where cues may be occluded, viewpoint-dependent, or transient. Appendix~\ref{app:real_results} gives all backbone results and per-concept accuracies.}

\textbf{Confidence describes relative support, not validated reliability.}
\revision{Mean normalized confidence correlates with rule accuracy across backbones (Spearman $\rho{=}0.92$ under the synthetic reference configuration and $0.64$ under the bee reference configuration, each across 12 backbones), but these aggregate associations do not establish calibration or individual-error detection. ConvNeXt~V2-L has mean confidence $0.967$ on bees despite only $85.8\%$ rule accuracy, illustrating that decisive scores can accompany errors. These results do not justify using an unvalidated confidence threshold to present explanations. We report complete results and additional analyses in Appendices~\ref{app:synthetic_results} and~\ref{app:real_results}.}

\section{Discussion, Limitations and Future Work}

\Fluent combines expert-authored logical rules, empirical label and concept prediction-error likelihoods, and visual examples of positively and negatively mentioned concepts. The selected rule interprets predicted evidence through the supplied vocabulary; it does not reconstruct the label classifier's internal reasoning. The results show that rule-implied labels can be more accurate than weak synthetic label heads: the three ResNets gain $26.7$--$46.3$ percentage points under the synthetic reference configuration. On bees, the direction reverses for every backbone, with gaps of $1.5$--$21.8$ points. These are descriptive comparisons under the evaluated data and estimation procedure. Mean rule confidence is associated with aggregate rule accuracy (Spearman $\rho=0.92$ for the synthetic reference configuration and $0.64$ for the bee reference configuration, each across 12 backbones), but this does not establish calibration or an error-detection threshold.

\textbf{Experimental limitations.}
\revision{The bee likelihood-estimation pool contains every explanation-evaluation image, concept training and test sources overlap, and observation families span the label splits. Its rule targets are class-derived templates without independent per-image concept verification. Inspected BatchNorm inference buffers differ from pretraining despite unchanged comparable learned parameters. One configured seed, incomplete training details, and uncontrolled possible background cues limit causal and generalization claims. Appendix~\ref{app:evaluation_limits} details these limitations.}

\textbf{Method and explanation limitations.}
The scorer assumes mutually exclusive rules covering the modeled domain and does not detect uncovered inputs. Its factorized error channels and prior are approximations. Candidate-specific evidence subsets make the normalized score different from a posterior conditioned on all observations and can favor shorter rules. The evaluated renderer supports positive disjunctions and unit negative literals, not arbitrary signed CNF. Occlusion highlights are sensitive to patch size and do not certify concept presence or anatomical boundaries. 

\textbf{Future work.}
A new evaluation could run user studies then assess understanding and task performance against other explanation methods like saliency, free-text, LLM-based and no-explanation conditions. Segmentation methods such as SAM 3~\citep{sam3} are candidates for localization, but their effect requires measurement. Overlapping rules and further domains with independently annotated concepts remain extensions.

{
    \small
    \bibliographystyle{ieeenat_fullname}
    \bibliography{main}
}

\clearpage
\appendix
\makeatletter
\setlength{\@fptop}{0pt}
\setlength{\@dblfptop}{0pt}
\makeatother

\section{Additional Method Details and Explanation Example}
\label{app:method_details}

{This appendix describes concept-likelihood estimation, derives the full-concept reference posterior and its relation to the implemented score, and gives an example explanation tuple.}

\subsection{Estimating Concept-Prediction Likelihoods}
\label{app:concept_likelihoods}

{The conditional likelihood $\Pr(\hat c_i=h\mid c_i=t)$ describes how often concept predictor $i$ outputs $h$ when the true concept state is $t$. It differs from the image-specific presence score $\revision{\phi_{c_i}}(x)$: the former summarizes prediction behavior over labeled examples, whereas the latter is the predictor's score for one image.

For concept $c_i$, let $\mathcal D^{test}_{c_i}=\{(x_j,t_j)\}_{j=1}^{N_{c_i}}$ be the designated concept-labeled evaluation set, with $t_j=c_i(x_j)\in\{0,1\}$. In the experiments, the binary output is $\hat c_i(x_j)=\mathbf1[\revision{\phi_{c_i}}(x_j)\ge0.5]$. We count true and predicted states using
\begin{equation}
\label{eq:concept_heldout_counts}
n_i(t,h)=\sum_{j=1}^{N_{c_i}}\mathbf1[t_j=t]\,\mathbf1[\hat c_i(x_j)=h],
\qquad t,h\in\{0,1\}.
\end{equation}
For a true state represented in the evaluation set, the empirical estimate is
\begin{equation}
\label{eq:concept_heldout_likelihood}
\Pr(\hat c_i=h\mid c_i=t)\;\approx\;
\frac{n_i(t,h)}{n_i(t,0)+n_i(t,1)}.
\end{equation}
The denominator counts all evaluation examples with true state $t$. The numerator counts those for which the predictor outputs $h$. Thus, each row of Table~\ref{table_conditional_probs} is normalized separately. For example, $\Pr(\hat c_i=1\mid c_i=1)$ is estimated by $\mathrm{TP}/(\mathrm{TP}+\mathrm{FN})$, while $\Pr(\hat c_i=1\mid c_i=0)$ is estimated by $\mathrm{FP}/(\mathrm{TN}+\mathrm{FP})$.

For a complete numerical specification, an empty true-state row is assigned probability $0.5$ for each output, and estimated concept likelihoods are clipped to $[10^{-12},1-10^{-12}]$ without pseudocounts. {These conditional likelihoods supply the concept-prediction factors in Equation~(\ref{eq_rule_post_restricted}). The same clipping convention applies to concept-prevalence parameters.}

}

\begin{table}[!ht]

\centering
\caption{Conditional likelihoods for a binary concept predictor, estimated from a designated concept-labeled test set via the confusion matrix.}
\label{table_conditional_probs}
\setlength{\tabcolsep}{6pt}
\begin{tabular}{c|cc}
\toprule
 & $\hat c_i=0$ & $\hat c_i=1$ \\
\midrule
$c_i=0$ & $\Pr(\hat c_i=0 \mid c_i=0)$ & $\Pr(\hat c_i=1 \mid c_i=0)$ \\
$c_i=1$ & $\Pr(\hat c_i=0 \mid c_i=1)$ & $\Pr(\hat c_i=1 \mid c_i=1)$ \\
\bottomrule
\end{tabular}
\end{table}

\subsection{{Full-Concept Posterior and Implemented Rule Score}}
\label{app:full_rule_posterior}

The coverage event $A=\{C\in\Omega\}$ restricts the reference model to concept assignments that satisfy a supplied rule. The independent product $q(\mathbf c)$ from Equation~(\ref{eq:pr_c}) is a base approximation before this restriction. The covered prior is
\[
\begin{gathered}
\Pr(C=\mathbf c\mid A)
=\frac{\mathbf1[\mathbf c\in\Omega]\,q(\mathbf c)}{Z_A},
\\
Z_A=\sum_{\mathbf u\in\Omega}q(\mathbf u)>0.
\end{gathered}
\]
Its concept marginals need not equal the fitted parameters $\pi_i$: conditioning on rule coverage can change their values and introduce dependence. This distinguishes the independent base approximation from the covered domain model.

Using the dependencies in Figure~\ref{fig_bayesnet}, the rule posterior given all predictions is
\begin{equation}
\label{eq_bayesnet}
\begin{aligned}
\Pr(r\mid\hat{\mathbf C},\hat\ell,A)
={}&\alpha_{\mathrm{full}}\sum_{l\in\mathcal L}\sum_{\mathbf c\in\Omega}
\Pr(\hat\ell\mid l)\Pr(l\mid r)\mathbf1[\mathbf c\models r]\\[-2pt]
&\hspace{22mm}\cdot\Pr(\hat{\mathbf C}\mid\mathbf c)q(\mathbf c).
\end{aligned}
\end{equation}
The normalizer $\alpha_{\mathrm{full}}$ absorbs the common coverage constant $Z_A$ and makes the posterior sum to one over rules. Because $\Pr(l\mid r)$ is zero for every $l\ne\ell(r)$, only $l=\ell(r)$ survives the class-label sum. Every assignment satisfying $r$ is already in $\Omega$, so
\begin{multline}
\label{eq_bayesnet_reduced}
\Pr(r\mid\hat{\mathbf C},\hat\ell,A)\\
=\alpha_{\mathrm{full}}\underbrace{\Pr(\hat\ell\mid\ell(r))
\sum_{\mathbf c:\mathbf c\models r}\prod_{i=1}^k\Pr(\hat c_i\mid c_i)\Pr(c_i)}_{w_{\mathrm{full}}(r,x)}.
\end{multline}
The corresponding full-evidence selection would be
\begin{equation}
\label{eq_r*x}
r^*_{\mathrm{full}}(x)
=\operatorname*{arg\,max}_{r\in\mathcal R_{expert}}\Pr(r\mid\hat{\mathbf C}(x),\hat\ell(x),A).
\end{equation}
This is a reference decision rule; the reported experiments use $r^*(x)$ from Equation~(\ref{eq_r*_final}).

To compare the two scores, define
$ a_i(t)=\Pr(\hat c_i\mid c_i=t)\Pr(c_i=t)$ and
$m_i(x)=\sum_{t\in\{0,1\}}a_i(t)$.
Since satisfaction of $r$ depends only on concepts in $\mathcal I_{\rho}(r)$, summing over the other true concept states gives
\begin{equation}
\label{eq:full_restricted_relation}
w_{\mathrm{full}}(r,x)=w(r,x)\prod_{i\notin\mathcal I_{\rho}(r)}m_i(x).
\end{equation}
Each omitted factor is the marginal likelihood of an observed concept prediction under the independent base model. The product generally depends on the candidate rule, so it cannot be absorbed into a shared normalization constant. Consequently, the full posterior and the implemented score can rank rules differently.

Conditional independence allows this factorization; it does not make these observed likelihoods disappear. For a concept not mentioned by the rule, if its prediction were unobserved, summing over both its predicted value and its true state would contribute $\sum_t\Pr(c_i=t)\sum_h\Pr(\hat c_i=h\mid c_i=t)=1$. The implemented procedure instead chooses which predictions to use separately for each candidate. Its confidence therefore expresses normalized relative support under that scoring choice. In particular, if every concept prediction is uninformative with likelihood $1/2$, the restricted score contains a factor $2^{-|\mathcal I_{\rho}(r)|}$ multiplying the rule's base-prior satisfaction mass and label likelihood. Rules mentioning different numbers of concepts can therefore receive different evidence-size penalties.

Equation~(\ref{eq_r*_final}) assumes a positive total weight. The numerical convention for zero total weight assigns the uniform value $1/|\mathcal R_{expert}|$ and selects the first candidate. Ties at the maximum weight are likewise resolved by candidate order. This fallback is a numerical convention, not additional evidence for the selected rule.

\clearpage
\twocolumn[{%
\subsection{A Complete Synthetic Explanation}
\label{app:explanation_example}
\begin{center}
  \includegraphics[width=0.62\textwidth,trim=0 15bp 0 23bp,clip]{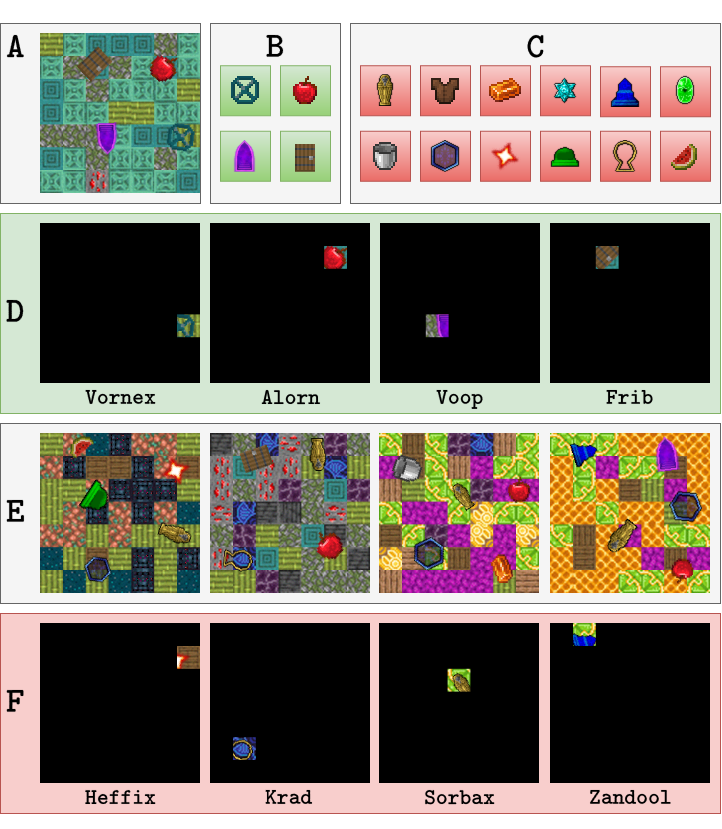}
    \captionsetup{hypcap=false}
    \captionof{figure}{{An illustrative \textsc{Fluent} display for a synthetic \textit{Kallow} image. This figure illustrates the explanation format; it does not establish a numerical confidence or nearest-reference ranking.} \textbf{(A)}~The input image containing multiple concept objects on a tiled background. \textbf{(B)}~The four concepts predicted as present by the concept predictor: $\mathtt{Alorn}$, $\mathtt{Vornex}$, $\mathtt{Voop}$, and $\mathtt{Frib}$. \textbf{(C)}~The twelve concepts predicted as absent. Rule: $\lnot \mathtt{Heffix} \land \lnot \mathtt{Krad} \land \lnot \mathtt{Sorbax} \land \lnot \mathtt{Zandool} \land (\mathtt{Alorn} \lor \mathtt{Vornex}) \land (\mathtt{Frib} \lor \mathtt{Voop})$ is provided as the best explanation. {\textbf{(D)}~Displayed query highlights for the positively mentioned concepts. These panels illustrate the output format; they are not a quantitative localization validation.} {\textbf{(E)}~Reference images containing each negatively mentioned concept.} {\textbf{(F)}~Occlusion highlights for these concepts in the reference images.}}
  \label{fig_FLUENT}
    \captionsetup{hypcap=true}
\end{center}
}]

\clearpage

\section{Datasets, Expert Rules, and Splits}
\label{app:datasets}

We evaluate \Fluent on a synthetic dataset and a real-world bee dataset. Each dataset is described below; Appendix~\ref{app:protocol} presents the experimental protocol and evaluation metrics, and Section~\ref{sec:results} summarizes the main results.

\subsection{Synthetic Dataset}
\label{sec:synthetic_dataset}

We evaluate \Fluent on a purpose-built synthetic dataset in which ground-truth labels, concepts, and rules are known. This controlled setting lets us evaluate label predictions, concept predictions, and rule selection against known ground truth, without the annotation uncertainty of natural images.

The synthetic dataset comprises $|\mathcal{L}|=3$ classes, $|\mathcal{C}|=16$ abstract concepts, and $|\mathcal{R}_{expert}|=6$ expert rules (two per class), listed in Table~\ref{tab:synthetic_rules}. Each concept is rendered as a distinct synthetic visual pattern, and images are composed by placing concept instances at random positions and scales on a tiled $224 \times 224$ background. Every ground-truth concept vector satisfies exactly its target rule. Exhaustive Boolean enumeration verifies pairwise mutual exclusivity (Section~\ref{sec:labels_concepts_rules}); coverage concerns the generated images, rather than every Boolean concept assignment (Equation~(\ref{eq_unique_rule})). Each rule mentions 8 of the 16 concepts. Of the 16 concepts, 12 participate in at least one rule and 4 ($\mathtt{Floom}$, $\mathtt{Opple}$, $\mathtt{Yintar}$, and $\mathtt{Zorf}$) appear in images but are irrelevant to classification.

Background choices and the generating software version are unspecified, and the evaluation does not isolate possible background shortcuts. Moreover, changing $T$ changes the evaluation images as well as the training pool: for example, the Vumble rule-1 count is 33, 44, and 62 out of 100 Explanation Test images at $T=500$, $2000$, and $5000$, respectively. Cross-$T$ results therefore combine training-size and data-composition differences.

\begin{table}[!ht]

\centering
\caption{CNF rules for the synthetic dataset (two rules per class). Each rule consists of two disjunctive clauses and four negated concepts.}
\label{tab:synthetic_rules}
\setlength{\tabcolsep}{4pt}
\footnotesize
\begin{tabular}{l l}
\toprule
\textbf{Class} & \textbf{Expert Rule} \\
\midrule
Kallow & $(\mathtt{Alorn}\vee\mathtt{Vornex})\wedge(\mathtt{Krad}\vee\mathtt{Zandool})$ \\
 & $\quad\wedge \lnot\mathtt{Guffit}\wedge\lnot\mathtt{Smeek}\wedge\lnot\mathtt{Hebda}\wedge\lnot\mathtt{Wexil}$ \\
\addlinespace[2pt]
Kallow & $(\mathtt{Alorn}\vee\mathtt{Vornex})\wedge(\mathtt{Frib}\vee\mathtt{Voop})$ \\
 & $\quad\wedge \lnot\mathtt{Krad}\wedge\lnot\mathtt{Zandool}\wedge\lnot\mathtt{Heffix}\wedge\lnot\mathtt{Sorbax}$ \\
\addlinespace[2pt]
Vumble & $(\mathtt{Guffit}\vee\mathtt{Smeek})\wedge(\mathtt{Frib}\vee\mathtt{Voop})$ \\
 & $\quad\wedge \lnot\mathtt{Alorn}\wedge\lnot\mathtt{Vornex}\wedge\lnot\mathtt{Heffix}\wedge\lnot\mathtt{Sorbax}$ \\
\addlinespace[2pt]
Vumble & $(\mathtt{Guffit}\vee\mathtt{Smeek})\wedge(\mathtt{Heffix}\vee\mathtt{Sorbax})$ \\
 & $\quad\wedge \lnot\mathtt{Frib}\wedge\lnot\mathtt{Voop}\wedge\lnot\mathtt{Krad}\wedge\lnot\mathtt{Zandool}$ \\
\addlinespace[2pt]
Dadron & $(\mathtt{Hebda}\vee\mathtt{Wexil})\wedge(\mathtt{Heffix}\vee\mathtt{Sorbax})$ \\
 & $\quad\wedge \lnot\mathtt{Alorn}\wedge\lnot\mathtt{Vornex}\wedge\lnot\mathtt{Guffit}\wedge\lnot\mathtt{Smeek}$ \\
\addlinespace[2pt]
Dadron & $(\mathtt{Hebda}\vee\mathtt{Wexil})\wedge(\mathtt{Krad}\vee\mathtt{Zandool})$ \\
 & $\quad\wedge \lnot\mathtt{Guffit}\wedge\lnot\mathtt{Smeek}\wedge\lnot\mathtt{Heffix}\wedge\lnot\mathtt{Sorbax}$ \\
\bottomrule
\end{tabular}
\end{table}

\begin{table}[!ht]

\centering
\caption{\revision{Data splits used in the synthetic reference configuration ($T{=}5000$, $C{=}500$, $P{=}0.1$). The \emph{Explanation Test} split provides ground-truth concept annotations and is used exclusively for evaluating the \Fluent rule selector (Section~\ref{sec:rule_selection}). The \emph{Contrastive} split provides images for retrieval of negated concepts.}}
\label{tab:data_splits}
\setlength{\tabcolsep}{2pt}
\footnotesize
\begin{tabular}{l r r l}
\toprule
\textbf{Split} & \textbf{Img./Class} & \textbf{Total} & \textbf{Used by} \\
\midrule
Label Train      & 5\,000 & 15\,000 & $\revision{\phi_{\mathrm{cls}}}$ training \\
Label Test       & 1\,000 & 3\,000 & $\revision{\phi_{\mathrm{cls}}}$ evaluation \\
Explanation Test & 100    & 300    & \Fluent benchmark \\
\midrule
\multicolumn{1}{l}{} & \textbf{Img./Concept} & \textbf{Total} & \\
\midrule
Concept Train & 500 pos./neg. pairs & 16\,000 & $\revision{\phi_{c_i}}$ training \\
Concept Test  & 100 pos./neg. pairs & 3\,200 & $\revision{\phi_{c_i}}$ evaluation \\
Contrastive Examples               & 100             & 1600    & $\mathcal{H}^-$ highlights\\
\bottomrule
\end{tabular}
\end{table}

\clearpage
\subsection{Real Bee Dataset}
\label{sec:real_dataset}

The bee benchmark contains field photographs grouped into \emph{Bumble Bee}, \emph{Green Metallic Sweat Bee}, and \emph{Hairy Leg Bee}. The six species are \emph{Bombus griseocollis} and \emph{Bombus huntii} (Bumble Bee), \emph{Agapostemon virescens} and \emph{Augochlorella aurata} (Green Metallic Sweat Bee), and \emph{Anthophora urbana} and \emph{Melissodes bimaculatus} (Hairy Leg Bee). The 600-image Explanation Test split contains 100 images per species and 200 per group.

The benchmark concept vectors are constant within each group: Bumble Bee has only $\mathtt{FuzzyBody}$; Green Metallic Sweat Bee has $\mathtt{PollenHindLeg}$ and $\mathtt{GreenHeadAndThorax}$; Hairy Leg Bee has $\mathtt{FuzzyBody}$ and $\mathtt{DenseHindLeg}$. Prior estimation uses the same species-to-concept templates. Independent human verification of which cues are visible in each photograph is not established, and complete source-image attribution and collection details are unavailable. We therefore evaluate agreement with these class-associated templates; these labels do not establish independently annotated image-visible concepts. The concept-classifier test sets are separate curated positive/negative entries and have the overlap limitations described below.

The four concepts used in the experiment are \emph{fuzzy hair on the entire body} ($\mathtt{FuzzyBody}$), \emph{dense hair on the hind legs} ($\mathtt{DenseHindLeg}$), \emph{pollen carried on the hind legs} ($\mathtt{PollenHindLeg}$), and \emph{a green head and thorax} ($\mathtt{GreenHeadAndThorax}$). The supplied rule set has one CNF rule per class, shown in Table~\ref{tab:real_rules}; these are task-specific rules, not universal species-identification criteria.

\begin{table}[!ht]

\centering
\caption{CNF rules for the bee benchmark (one rule per class).}
\label{tab:real_rules}
\setlength{\tabcolsep}{4pt}
\footnotesize
\begin{tabular}{l l}
\toprule
\textbf{Class} & \textbf{Expert Rule} \\
\midrule
Bumble Bee & $\mathtt{FuzzyBody} \wedge \lnot\mathtt{DenseHindLeg}$ \\
\addlinespace[2pt]
Green Metallic & $\mathtt{PollenHindLeg} \wedge \lnot\mathtt{FuzzyBody}$ \\
Sweat Bee & $\;\wedge\; \mathtt{GreenHeadAndThorax} \wedge \lnot\mathtt{DenseHindLeg}$ \\
\addlinespace[2pt]
Hairy Leg Bee & $\mathtt{FuzzyBody} \wedge \mathtt{DenseHindLeg}$ \\
\bottomrule
\end{tabular}
\end{table}

The three rules are pairwise mutually exclusive and cover 9 of the 16 Boolean assignments; each benchmark template satisfies its group's rule. They are not exhaustive over all Boolean concept assignments, consistent with the restricted coverage assumption in Equation~(\ref{eq_unique_rule}). Table~\ref{tab:real_splits} gives the dataset counts. Concept entries are counted separately for each concept (positives plus negatives), so their totals need not equal the number of distinct source photographs. The intended augmentation factor is $8\times$, but the augmented training images total 32\,928, rather than $4\,119\times8=32\,952$; three pairs of source files share a filename stem, consistent with the 24-entry shortfall.

\begin{table}[!ht]

\centering
\caption{Counts for the bee experiment. ``Test'' names identify the dataset partitions; they do not imply independent evaluation (see the overlap analysis below). The outer Label Train count includes 3\,643 validation entries. Contrastive counts exclude three non-image metadata files.}
\label{tab:real_splits}
\setlength{\tabcolsep}{2pt}
\footnotesize
\begin{tabular}{l r r l}
\toprule
\textbf{Split} & \textbf{Img./Class} & \textbf{Total} & \textbf{Used by} \\
\midrule
Label Train      & {24\,271--24\,301} & {72\,873} & $\revision{\phi_{\mathrm{cls}}}$ training \\
Label Test       & {2\,698--2\,729} & {8\,125} & $\revision{\phi_{\mathrm{cls}}}$ evaluation \\
Explanation Test & {200} & {600} & \Fluent benchmark \\
\midrule
\multicolumn{1}{l}{} & \textbf{Img./Concept} & \textbf{Total} & \\
\midrule
Concept Train        & {929--1\,160} & {4\,119} & $\revision{\phi_{c_i}}$ training \\
Concept Test         & {118--184} & {577} & $\revision{\phi_{c_i}}$ evaluation \\
Contrastive Examples & {186--422} & {1\,236} & $\mathcal{H}^-$ highlights \\
\bottomrule
\end{tabular}
\end{table}

\paragraph{Split overlap and evaluation limitations.}
All 600 Explanation Test images have byte-identical SHA-256 matches in the 8\,125-image Label Test partition used to estimate the label-confusion likelihood. Thus the evaluation examples contribute to a fitted component of the rule scorer. Removing those exact images would leave 7\,525 calibration images, but the reported results use the original likelihoods. Additionally, 447 of the 600 benchmark images and 6\,191 of the 8\,125 Label Test images share a filename-derived observation family with a Label Train image. These family counts indicate sibling-view overlap; they do not quantify its effect on accuracy.

For the concept classifiers, 21 of the 577 Concept Test entries (20 distinct image hashes) match a training source image for the same concept: 3 fuzzy-body, 8 dense-hind-leg, 1 pollen-hind-leg, and 9 green-head/thorax entries. Their test supports are respectively 134, 184, 118, and 141. Training uses augmented sources, so this is source-image overlap even when the final augmented tensors differ. Exact hashing does not detect all crops, re-encodings, or related views. Consequently, the accuracies and likelihood estimates are not independent at the source-image or observation level. Independent evaluation requires grouped, deduplicated splits and new fitting and scoring.

\paragraph{Evaluation scope.}
The concept configuration is labeled ``100 examples''; the actual counts used in the experiment are given in Table~\ref{tab:real_splits}. This paper reports automatic evaluation only and no completed human study.

\clearpage
\section{Experimental Protocol and Metrics}
\label{app:protocol}

\subsection{Backbones and Training Settings}
\label{sec:experimental_setup}

\revision{We evaluate the 12 \texttt{timm} backbones~\citep{timm} in Table~\ref{tab:backbone_models}. They include ConvNeXt V2~\citep{ConvNeXtV2}, ResNets with torchvision or RSB recipes~\citep{resnet,rsb}, DINO~\citep{vit}, MAE~\citep{mae}, SigLIP 2~\citep{siglip2}, BEiT~\citep{beit}, Swin~\citep{swinv1}, and SwinV2~\citep{swin}. These references identify architecture or pretraining families; the model identifiers specify the actual variants. SigLIP 2 specifies WebLI.}

\begin{table*}[tp]
\centering
\caption{Backbone identifiers and concept-probe layers. CNXv2-L without a qualifier means the fine-tuned variant; RN152 means A1h.}
\label{tab:backbone_models}
\scriptsize
\setlength{\tabcolsep}{3pt}
\begin{tabular}{p{0.22\linewidth}p{0.53\linewidth}p{0.20\linewidth}}
\toprule
Paper label & Model identifier & Probe layer \\
\midrule
CNXv2-L & \path{convnextv2_large.fcmae_ft_in22k_in1k} & \path{head.global_pool} \\
RN152 (A1h) & \path{resnet152.a1h_in1k} & \path{global_pool} \\
ViT-B/16 (DINO) & \path{vit_base_patch16_224.dino} & \path{fc_norm} \\
SwinV2-S & \path{swinv2_cr_small_ns_224.sw_in1k} & \path{head.global_pool} \\
ViT-B/16 (SigLIP 2) & \path{vit_base_patch16_siglip_gap_224.v2_webli} & \path{fc_norm} \\
BEiT-L/16 & \path{beit_large_patch16_224.in22k_ft_in22k_in1k} & \path{fc_norm} \\
ViT-B/16 (MAE) & \path{vit_base_patch16_224.mae} & \path{fc_norm} \\
CNXv2-L (FCMAE) & \path{convnextv2_large.fcmae} & \path{head.global_pool} \\
RN50 (TV) & \path{resnet50.tv_in1k} & \path{global_pool} \\
RN50 (A1h) & \path{resnet50.a1h_in1k} & \path{global_pool} \\
Swin-B (IN-1k) & \path{swin_base_patch4_window7_224.ms_in1k} & \path{head.global_pool} \\
Swin-B (IN-22k) & \path{swin_base_patch4_window7_224.ms_in22k} & \path{head.global_pool} \\
\bottomrule
\end{tabular}
\end{table*}

\revision{The synthetic settings use $T\in\{500,2000,5000\}$ label-training images per class, $C\in\{100,200,500\}$ positive and equally many negative concept examples, $P\in\{0.1,0.2\}$ prior sample fractions, and configured seed 42. This gives 216 model/configuration results, or 108 at either prior fraction. Both datasets use linear concept heads, no layer search, augmentation setting 8, and a configured label-head limit of 25 epochs. The synthetic configuration additionally specifies a concept-head limit of 100 epochs. These are epoch limits; individual runs need not reach them. Optimizer and learning-rate settings, complete training histories, exact software revisions, and checkpoint-to-result links are not fully documented, limiting exact reproducibility.}

\subsection{Backbone Inference State}
\label{app:backbone_state}

\revision{The stated training design fixes learned backbone parameters and trains task heads, but this does not imply an unchanged feature mapping. Comparing the RN50-TV and RN152-A1h checkpoints with cached pretrained weights finds all comparable non-head learned parameters unchanged, while all inspected BatchNorm running-statistic and counter buffers differ: 159/159 and 465/465, respectively, in each dataset. Such buffers affect inference even without parameter-gradient updates. This comparison covers these two available checkpoint pairs per dataset and does not establish the state of every checkpoint. We consequently avoid attributing cross-backbone differences solely to unchanged pretrained representations.}

\subsection{Likelihoods and Prior Estimation}
\label{sec:real_setup}

The label likelihood $\Pr(\hat\ell\mid\ell)$ uses the Label Test confusion matrix, with rows conditioned on the true class. Each binary concept likelihood uses the Concept Test counts as in Appendix~\ref{app:concept_likelihoods}. ConceptAcc is an unweighted mean of the concept-specific accuracies, rather than accuracy pooled across differently sized concept sets. Label and rule percentages are computed from their integer correct counts.

\revision{Synthetic concept priors use generated concept metadata sampled from Label Train. Bee priors use the species-to-concept mapping described in Appendix~\ref{sec:real_dataset}. The bee sample sizes are 7\,290 at $P=0.1$ and 14\,577 at $P=0.2$. For the reference-protocol comparison, the two resulting priors differ slightly but all rounded bee metrics agree. The ConvNeXt V2-L label head has $95.4\%$ accuracy on 8\,125 Label Test images; the other heads range from $60.1\%$ to $94.7\%$. This larger pool contains the 600-image Explanation Test used for Explanation Test LabelAcc.}

\subsection{Evaluation Limitations}
\label{app:evaluation_limits}

\paragraph{Bee data independence.}
All 600 Explanation Test images have byte-identical copies in the 8\,125-image Label Test pool used to estimate the label likelihoods. Removing these copies would leave 7\,525 images, but no corrected result is inferred from that subtraction. The Concept Test set includes 21 of 577 entries duplicating a training source for the same concept. Filename-based observation-family grouping also finds training siblings for 447/600 explanation images and 6\,191/8\,125 Label Test images. These counts distinguish exact-file overlap from related views of the same observation. The scores therefore do not establish generalization to independent observations or independent likelihood-estimation data. The nominal 72\,873-image Label Train pool has a 3\,643-image internal validation set; its complement contains 69\,230 images if that partition is exhaustive.

\paragraph{Scope of comparison.}
The experiments use one configured seed and do not estimate uncertainty across training runs. Sweep cells share models and data, and are not independent replications. Changing $T$ also changes the synthetic evaluation images and rule proportions; it is not a pure training-size intervention. Background shortcuts are not separately controlled. The evaluation does not compare explanation methods, ablate individual components, assess confidence calibration, quantify localization accuracy, or report user-study outcomes. The qualitative examples illustrate the display format without validating localization or reference retrieval.

\subsection{Evaluation Metrics}
\label{sec:eval_metrics}

For each Explanation Test image, let $r^{gt}(x)$ be the target rule implied by the concept vector and $\ell^{gt}(x)$ its label. Synthetic vectors come from generation metadata; bee vectors are the class-associated templates described in Appendix~\ref{sec:real_dataset}.

\textbf{LabelAcc} is the fraction with $\hat\ell(x)=\ell^{gt}(x)$. \textbf{ConceptAcc} is the mean binary accuracy over 16 synthetic or four bee Concept Test reports, not accuracy on Explanation Test concepts. \textbf{RuleAcc} is the fraction with $r^*(x)=r^{gt}(x)$. \textbf{RuleLabelAcc} is the fraction with $\ell(r^*(x))=\ell^{gt}(x)$. It can exceed LabelAcc without changing the classifier output. RuleAcc cannot exceed RuleLabelAcc, and they are equal for the one-rule-per-class bee benchmark. \textbf{Avg~$\operatorname{Conf}$} averages the selected rule's normalized score; it measures relative decisiveness, not calibrated correctness.

\subsection{\revision{Results Collection}}
\label{app:results_collection}

\revision{The separately provided \path{experiment_results/} directory contains the complete numerical results, copied source reports, and file hashes. Its \path{benchmark_results.csv} gives all five aggregate metrics at full precision for every included configuration, with explicit dataset, model, training-budget, prior-fraction, and protocol fields. Per-concept metrics, label-head evaluations, and prior estimates are provided separately. The appendix prints all aggregate configurations, per-concept accuracies, and label-head summaries; the CSV and JSON files retain the underlying counts and additional metric fields.}

\revision{The reference comparison contains 216 synthetic configurations and 24 bee configurations across 12 backbones.}

\clearpage
\section{Complete Synthetic Results and Sensitivity Analyses}
\label{app:results}
\label{app:synthetic_results}

\revision{This appendix reports all 216 synthetic configurations: 12 backbones, $T\in\{500,2000,5000\}$ label-training images per class, $C\in\{100,200,500\}$ positive and the same number of negative training examples per concept, and prior-estimation fractions $P\in\{0.1,0.2\}$. The \emph{synthetic reference configuration} used in the main paper is $T{=}5000$, $C{=}500$, $P{=}0.1$. Each configuration is evaluated once; repeated-seed uncertainty estimates are unavailable. Appendix~\ref{app:protocol} defines the metrics and training protocol.}

\revision{Benchmark accuracy percentages are rounded half-up to one decimal, and mean confidence to three decimals. LabelAcc, RuleAcc, RuleLabelAcc, and Avg~$\operatorname{Conf}$ use the 300-image Explanation Test split; ConceptAcc is the unweighted mean of 16 binary concept-test accuracies. Each concept test has 100 positive and 100 negative examples. The separately provided \texttt{experiment\_results} directory contains full-precision metrics, correct-prediction counts, concept confusion matrices, classification reports, and the source reports used to generate these tables. Tables in this appendix present all five benchmark metrics for every configuration and the individual concept and label-head accuracies. The complete sweep tables also give Switch\%, the fraction of images for which the selected rule implies a different class from the classifier prediction; a switch is not necessarily an error.}

\subsection{\revision{Synthetic reference configuration}}

\begin{table*}[tp]
\centering
\caption{\revision{All 12 backbones in the synthetic reference configuration ($T{=}5000$, $C{=}500$, $P{=}0.1$). Accuracies are percentages and confidence is unitless. Bold marks the maximum in each column.}}
\label{tab:main_results}
\setlength{\tabcolsep}{3pt}
\footnotesize
\begin{tabular}{lccccc}
\toprule
\textbf{Backbone} & \revision{\textbf{LabelAcc}} & \revision{\textbf{ConceptAcc}} & \revision{\textbf{RuleAcc}} & \revision{\textbf{RuleLabelAcc}} & \revision{\textbf{Avg Conf}} \\
\midrule
CNXv2-L & 99.3\% & 93.3\% & 93.7\% & \textbf{99.7\%} & 0.925 \\
CNXv2-L (FCMAE only) & 94.7\% & 56.9\% & 54.0\% & 94.0\% & 0.687 \\
RN152 & 40.3\% & 73.6\% & 47.7\% & 67.3\% & 0.540 \\
RN50 (TV) & 39.0\% & 79.9\% & 69.7\% & 85.3\% & 0.634 \\
RN50 (A1h) & 51.0\% & 87.0\% & 67.7\% & 77.7\% & 0.698 \\
ViT-B/16 (DINO) & \textbf{99.7\%} & 92.8\% & \textbf{96.7\%} & \textbf{99.7\%} & 0.935 \\
ViT-B/16 (MAE) & 97.7\% & 79.4\% & 75.0\% & 97.7\% & 0.804 \\
ViT-B/16 (SigLIP\,2) & 99.0\% & 91.6\% & 90.0\% & 99.0\% & 0.908 \\
BEiT-L/16 & 99.3\% & 89.4\% & 91.3\% & \textbf{99.7\%} & 0.938 \\
Swin-B (IN-1k) & 99.0\% & 90.0\% & 90.7\% & 99.0\% & 0.918 \\
Swin-B (IN-22k) & 99.0\% & \textbf{95.2\%} & 94.7\% & 99.0\% & \textbf{0.949} \\
SwinV2-S & 99.3\% & 91.9\% & 94.0\% & 99.3\% & 0.918 \\
\bottomrule
\end{tabular}
\end{table*}

\subsection{Detailed synthetic findings}

\revision{{\emph{Rule-implied labels can correct classifier errors.}
The three ResNet backbones illustrate this property dramatically: their transfer-trained heads reach only $39.0$--$51.0\%$ label accuracy on the Rule Selector Test split (RN50~(TV) $39.0\%$, RN152 $40.3\%$, RN50~(A1h) $51.0\%$), yet the rule-implied label accuracy (RuleLabelAcc) reaches $85.3\%$, $67.3\%$, and $77.7\%$, respectively --- improvements of $+46.3$, $+27.0$, and $+26.7$ points obtained by incorporating concept evidence into the joint rule score. This occurs because the score in Equation~(\ref{eq_rule_post_restricted}) can rank a rule of another class above rules associated with the classifier prediction $\hat\ell(x)$. The concept classifiers, while individually imperfect ($73.6$--$87.0\%$ mean accuracy for the three ResNets), collectively provide enough evidence for the rule selector to infer a different label from an incorrect classifier prediction. The correction is positive in every $(T, C)$ cell of the sweep at $P{=}0.1$ for all three ResNets (RN50~(TV) $+6.7$ to $+54.3$ points, RN50~(A1h) $+15.3$ to $+29.3$, RN152 $+13.7$ to $+27.0$; Appendix~\ref{app:full_sweep}). For the nine backbones whose heads are at least $90\%$ accurate in the synthetic reference configuration, rule-implied label accuracy remains close to classifier accuracy: RuleLabelAcc differs from LabelAcc by at most $0.7$ points in the synthetic reference configuration (\eg CNXv2-L $99.3\% \rightarrow 99.7\%$, CNXv2-L~(FCMAE only) $94.7\% \rightarrow 94.0\%$) and by between $-1.7$ and $+0.7$ points over the sweep at $P{=}0.1$. {More generally, each synthetic class has two mutually exclusive rules, and each image satisfies exactly one of them. RuleLabelAcc can exceed RuleAcc when the selector chooses the other rule of the correct class: the rule is incorrect for that image, but its implied label is correct.}}}

\revision{{\emph{\Fluent rule accuracy tracks concept quality, not just label accuracy.}
Across the 12 backbones in the synthetic reference configuration, rule accuracy is more strongly rank-correlated with mean concept accuracy (Spearman $\rho = 0.90$) than with label accuracy ($\rho = 0.87$); over their 108 (backbone, $T$, $C$) cells at $P{=}0.1$ the correlations are $\rho = 0.91$ and $0.79$, respectively. These are descriptive associations among related configurations, not independent observations or significance tests. Backbones with near-perfect heads but weak concept probes are the clearest cases: ViT-B/16~(MAE) reaches $97.7\%$ label accuracy but only $79.4\%$ mean concept accuracy and $75.0\%$ rule accuracy, and CNXv2-L~(FCMAE only) reaches $94.7\%$ label accuracy but $56.9\%$ concept and $54.0\%$ rule accuracy; conversely, RN50~(TV) attains $69.7\%$ rule accuracy from a $39.0\%$ head because its concept probes average $79.9\%$. The same effect is visible within a single backbone as the concept-annotation budget grows: at $T{=}5000,P{=}0.1$, raising $C$ from $100$ to $200$ to $500$ increases mean concept accuracy monotonically for all 12 backbones, and rule accuracy rises from $C{=}100$ to $C{=}500$ for all 12 of them (strictly at every step for 10; ViT-B/16~(SigLIP\,2) is flat at $85.0\%$ between $C{=}100$ and $C{=}200$, and BEiT-L/16 dips at $C{=}200$, $81.7\% \rightarrow 80.0\% \rightarrow 91.3\%$), \eg CNXv2-L $90.0\% \rightarrow 92.3\% \rightarrow 93.7\%$, SwinV2-S $87.0\% \rightarrow 93.3\% \rightarrow 94.0\%$, RN50~(TV) $35.3\% \rightarrow 41.7\% \rightarrow 69.7\%$, while label accuracy is unaffected because the head does not depend on $C$. Because each rule mentions $8$ concepts, uncertainty across their predictions compounds in the rule score, and rule selection is sensitive to per-concept errors.}}

{\emph{Pretraining variants differ in rule recovery.}
For related architecture families, the variants differ by tens of points in rule accuracy: ConvNeXt~V2-L with supervised fine-tuning reaches $93.7\%$ rule accuracy against $54.0\%$ for the same architecture with FCMAE pretraining only (mean concept accuracy $93.3\%$ vs.\ $56.9\%$); ViT-B/16 reaches $96.7\%$ with DINO, $90.0\%$ with SigLIP\,2, and $75.0\%$ with MAE weights; and Swin-B improves from $90.7\%$ to $94.7\%$ when pretrained on IN-22k instead of IN-1k. Label accuracy remains $94.7$--$99.7\%$ across these seven variants. These comparisons are consistent with an association between concept-prediction quality and rule recovery, but they do not isolate the effects of pretraining data, objectives, and implementation differences.}

\revision{{\emph{Average rule confidence correlates with rule accuracy.}
Across the 12 backbones in the synthetic reference configuration, mean normalized rule confidence correlates with rule accuracy with Spearman $\rho = 0.92$ (Pearson $r = 0.96$): every backbone with Avg~${\operatorname{Conf}} \geq 0.90$ attains at least $90.0\%$ rule accuracy (seven backbones, $90.0$--$96.7\%$), and every backbone below $0.90$ attains at most $75.0\%$ (five backbones, $47.7$--$75.0\%$). Over their 108 sweep cells at $P{=}0.1$ the rank correlation is $\rho = 0.95$. Within a single backbone the nine $(T,C)$ cells at $P{=}0.1$ span a narrower range and the relation is weaker ($\rho$ between $0.20$ and $0.93$ depending on the backbone), so the observed association is stronger across the evaluated configurations than within individual backbones. These aggregate associations do not establish that confidence reliably identifies individual explanation errors.}}

\clearpage
\subsection{Complete synthetic sweep}
\label{app:full_sweep}

\revision{Tables~\ref{tab:app_sweep_T500}--\ref{tab:app_sweep_T5000_P02} give every benchmark metric at both prior-estimation fractions. Label and concept predictors are shared between the two $P$ values, and label predictors are shared across $C$ at fixed $T$. Different $T$ settings use separately generated training and evaluation sets; consequently, changes across $T$ combine training-size and data-composition effects. For example, Vumble rule~1 appears in 163/500, 954/2000, and 2749/5000 training images and 33/100, 44/100, and 62/100 Explanation Test images, respectively. These comparisons therefore do not isolate sample efficiency.}

\revision{Table~\ref{tab:ablation_grid} summarizes rule accuracy at $P{=}0.1$ across all nine $(T,C)$ pairs.}

\begin{table*}[tp]
\centering
\caption{\revision{{Rule accuracy (\%) over the complete synthetic sweep at $P{=}0.1$: training images per class $T \in \{500, 2000, 5000\}$, positive and negative examples per concept $C \in \{100, 200, 500\}$, all 12 backbones. Every metric of every cell is listed in Appendix~\ref{app:full_sweep}.}}}
\label{tab:ablation_grid}
\setlength{\tabcolsep}{2pt}
\footnotesize
\begin{tabular}{l ccc ccc ccc}
\toprule
 & \multicolumn{3}{c}{{$T{=}500$}} & \multicolumn{3}{c}{{$T{=}2000$}} & \multicolumn{3}{c}{{$T{=}5000$}} \\
\cmidrule(lr){2-4}\cmidrule(lr){5-7}\cmidrule(lr){8-10}
{\textbf{Backbone}} & {$C{=}100$} & {$C{=}200$} & {$C{=}500$} & {$C{=}100$} & {$C{=}200$} & {$C{=}500$} & {$C{=}100$} & {$C{=}200$} & {$C{=}500$} \\
\midrule
{CNXv2-L} & {85.7} & {89.0} & {89.0} & {87.0} & {89.7} & {89.0} & {90.0} & {92.3} & {93.7} \\
{CNXv2-L (FCMAE only)} & {55.7} & {46.3} & {51.7} & {46.0} & {48.7} & {47.0} & {49.7} & {53.0} & {54.0} \\
{RN152} & {31.7} & {33.0} & {35.0} & {41.7} & {45.3} & {46.0} & {41.7} & {43.3} & {47.7} \\
{RN50 (TV)} & {45.7} & {59.0} & {62.3} & {42.3} & {52.3} & {73.0} & {35.3} & {41.7} & {69.7} \\
{RN50 (A1h)} & {53.7} & {57.7} & {63.7} & {55.3} & {61.7} & {69.7} & {53.7} & {59.0} & {67.7} \\
\midrule
{ViT-B/16 (DINO)} & {84.0} & {83.7} & {92.7} & {93.3} & {97.0} & {97.3} & {92.7} & {95.3} & {96.7} \\
{ViT-B/16 (MAE)} & {63.7} & {61.0} & {67.3} & {62.0} & {68.0} & {73.0} & {61.7} & {66.7} & {75.0} \\
{ViT-B/16 (SigLIP\,2)} & {79.3} & {82.7} & {89.7} & {75.7} & {85.0} & {87.7} & {85.0} & {85.0} & {90.0} \\
{BEiT-L/16} & {73.7} & {82.3} & {86.7} & {78.3} & {84.0} & {92.7} & {81.7} & {80.0} & {91.3} \\
\midrule
{Swin-B (IN-1k)} & {78.0} & {83.0} & {84.3} & {84.7} & {90.0} & {93.0} & {87.3} & {90.0} & {90.7} \\
{Swin-B (IN-22k)} & {87.7} & {89.0} & {91.0} & {87.3} & {92.3} & {94.7} & {91.0} & {94.0} & {94.7} \\
{SwinV2-S} & {75.3} & {79.7} & {83.3} & {81.7} & {83.7} & {92.7} & {87.0} & {93.3} & {94.0} \\
\bottomrule
\end{tabular}
\end{table*}

\begin{table*}[tp]
\centering
\caption{\revision{Complete synthetic results at $T{=}500$ and $P{=}0.1$: all 12 backbones and all three concept-training budgets. Accuracies and label-switch rates are percentages; confidence is unitless.}}
\label{tab:app_sweep_T500}
\setlength{\tabcolsep}{2pt}
\footnotesize
\begin{tabular}{lrcccccc}
\toprule
\textbf{Backbone} & $C$ & \revision{\textbf{LabelAcc}} & \revision{\textbf{ConceptAcc}} & \revision{\textbf{RuleAcc}} & \revision{\textbf{RuleLabelAcc}} & \revision{\textbf{Avg Conf}} & \revision{\textbf{Switch\%}} \\
\midrule
CNXv2-L & 100 & 99.3\% & 89.8\% & 85.7\% & 98.7\% & 0.903 & \revision{0.7} \\
 & 200 & 99.3\% & 91.7\% & 89.0\% & 99.7\% & 0.895 & \revision{0.3} \\
 & 500 & 99.3\% & 93.9\% & 89.0\% & 99.3\% & 0.916 & \revision{0.0} \\
\addlinespace[2pt]
CNXv2-L (FCMAE only) & 100 & 95.7\% & 55.2\% & 55.7\% & 95.3\% & 0.664 & \revision{1.0} \\
 & 200 & 95.7\% & 58.5\% & 46.3\% & 94.3\% & 0.681 & \revision{1.3} \\
 & 500 & 95.7\% & 57.8\% & 51.7\% & 95.7\% & 0.668 & \revision{0.3} \\
\addlinespace[2pt]
RN152 & 100 & 35.3\% & 67.1\% & 31.7\% & 49.0\% & 0.414 & \revision{51.0} \\
\bottomrule
\end{tabular}
\end{table*}

\clearpage

\begin{table*}[tp]
\centering
\caption{\revision{Complete synthetic results at $T{=}500$ and $P{=}0.2$: all 12 backbones and all three concept-training budgets. Accuracies and label-switch rates are percentages; confidence is unitless.}}
\label{tab:app_sweep_T500_P02}
\setlength{\tabcolsep}{2pt}
\footnotesize
\begin{tabular}{lrcccccc}
\toprule
\revision{\textbf{Backbone}} & \revision{$C$} & \revision{\textbf{LabelAcc}} & \revision{\textbf{ConceptAcc}} & \revision{\textbf{RuleAcc}} & \revision{\textbf{RuleLabelAcc}} & \revision{\textbf{Avg Conf}} & \revision{\textbf{Switch\%}} \\
\midrule
\revision{CNXv2-L} & \revision{100} & \revision{99.3\%} & \revision{89.8\%} & \revision{85.7\%} & \revision{98.7\%} & \revision{0.902} & \revision{0.7} \\
 & \revision{200} & \revision{99.3\%} & \revision{91.7\%} & \revision{89.0\%} & \revision{99.7\%} & \revision{0.895} & \revision{0.3} \\
 & \revision{500} & \revision{99.3\%} & \revision{93.9\%} & \revision{89.0\%} & \revision{99.3\%} & \revision{0.912} & \revision{0.0} \\
\addlinespace[2pt]
\revision{CNXv2-L (FCMAE only)} & \revision{100} & \revision{95.7\%} & \revision{55.2\%} & \revision{56.3\%} & \revision{95.3\%} & \revision{0.672} & \revision{1.0} \\
 & \revision{200} & \revision{95.7\%} & \revision{58.5\%} & \revision{48.0\%} & \revision{94.3\%} & \revision{0.686} & \revision{1.3} \\
 & \revision{500} & \revision{95.7\%} & \revision{57.8\%} & \revision{52.7\%} & \revision{95.7\%} & \revision{0.672} & \revision{0.3} \\
\addlinespace[2pt]
\revision{RN152} & \revision{100} & \revision{35.3\%} & \revision{67.1\%} & \revision{31.3\%} & \revision{49.3\%} & \revision{0.410} & \revision{51.0} \\
\bottomrule
\end{tabular}
\end{table*}

\clearpage

\begin{table*}[tp]
\centering
\caption{\revision{Complete synthetic results at $T{=}2000$ and $P{=}0.1$: all 12 backbones and all three concept-training budgets. Accuracies and label-switch rates are percentages; confidence is unitless.}}
\label{tab:app_sweep_T2000}
\setlength{\tabcolsep}{2pt}
\footnotesize
\begin{tabular}{lrcccccc}
\toprule
\textbf{Backbone} & $C$ & \revision{\textbf{LabelAcc}} & \revision{\textbf{ConceptAcc}} & \revision{\textbf{RuleAcc}} & \revision{\textbf{RuleLabelAcc}} & \revision{\textbf{Avg Conf}} & \revision{\textbf{Switch\%}} \\
\midrule
CNXv2-L & 100 & 99.7\% & 90.3\% & 87.0\% & 99.7\% & 0.901 & \revision{0.0} \\
 & 200 & 99.7\% & 91.1\% & 89.7\% & 99.7\% & 0.902 & \revision{0.0} \\
 & 500 & 99.7\% & 94.1\% & 89.0\% & 99.3\% & 0.917 & \revision{0.3} \\
\addlinespace[2pt]
CNXv2-L (FCMAE only) & 100 & 95.0\% & 55.1\% & 46.0\% & 93.3\% & 0.613 & \revision{1.7} \\
 & 200 & 95.0\% & 57.1\% & 48.7\% & 95.0\% & 0.723 & \revision{0.0} \\
 & 500 & 95.0\% & 57.3\% & 47.0\% & 94.3\% & 0.683 & \revision{0.7} \\
\addlinespace[2pt]
RN152 & 100 & 44.7\% & 67.0\% & 41.7\% & 61.3\% & 0.461 & \revision{42.3} \\
\bottomrule
\end{tabular}
\end{table*}

\clearpage

\begin{table*}[tp]
\centering
\caption{\revision{Complete synthetic results at $T{=}2000$ and $P{=}0.2$: all 12 backbones and all three concept-training budgets. Accuracies and label-switch rates are percentages; confidence is unitless.}}
\label{tab:app_sweep_T2000_P02}
\setlength{\tabcolsep}{2pt}
\footnotesize
\begin{tabular}{lrcccccc}
\toprule
\revision{\textbf{Backbone}} & \revision{$C$} & \revision{\textbf{LabelAcc}} & \revision{\textbf{ConceptAcc}} & \revision{\textbf{RuleAcc}} & \revision{\textbf{RuleLabelAcc}} & \revision{\textbf{Avg Conf}} & \revision{\textbf{Switch\%}} \\
\midrule
\revision{CNXv2-L} & \revision{100} & \revision{99.7\%} & \revision{90.3\%} & \revision{87.0\%} & \revision{99.7\%} & \revision{0.898} & \revision{0.0} \\
 & \revision{200} & \revision{99.7\%} & \revision{91.1\%} & \revision{90.0\%} & \revision{99.7\%} & \revision{0.899} & \revision{0.0} \\
 & \revision{500} & \revision{99.7\%} & \revision{94.1\%} & \revision{89.0\%} & \revision{99.3\%} & \revision{0.915} & \revision{0.3} \\
\addlinespace[2pt]
\revision{CNXv2-L (FCMAE only)} & \revision{100} & \revision{95.0\%} & \revision{55.1\%} & \revision{45.7\%} & \revision{93.3\%} & \revision{0.618} & \revision{1.7} \\
 & \revision{200} & \revision{95.0\%} & \revision{57.1\%} & \revision{49.0\%} & \revision{95.0\%} & \revision{0.731} & \revision{0.0} \\
 & \revision{500} & \revision{95.0\%} & \revision{57.3\%} & \revision{46.3\%} & \revision{94.3\%} & \revision{0.690} & \revision{0.7} \\
\addlinespace[2pt]
\revision{RN152} & \revision{100} & \revision{44.7\%} & \revision{67.0\%} & \revision{41.7\%} & \revision{61.3\%} & \revision{0.463} & \revision{42.3} \\
\bottomrule
\end{tabular}
\end{table*}

\clearpage

\begin{table*}[tp]
\centering
\caption{\revision{Complete synthetic results at $T{=}5000$ and $P{=}0.1$: all 12 backbones and all three concept-training budgets. Accuracies and label-switch rates are percentages; confidence is unitless.}}
\label{tab:app_sweep_T5000}
\setlength{\tabcolsep}{2pt}
\footnotesize
\begin{tabular}{lrcccccc}
\toprule
\textbf{Backbone} & $C$ & \revision{\textbf{LabelAcc}} & \revision{\textbf{ConceptAcc}} & \revision{\textbf{RuleAcc}} & \revision{\textbf{RuleLabelAcc}} & \revision{\textbf{Avg Conf}} & \revision{\textbf{Switch\%}} \\
\midrule
CNXv2-L & 100 & 99.3\% & 90.0\% & 90.0\% & 99.3\% & 0.908 & \revision{0.0} \\
 & 200 & 99.3\% & 91.6\% & 92.3\% & 99.7\% & 0.904 & \revision{0.3} \\
 & 500 & 99.3\% & 93.3\% & 93.7\% & 99.7\% & 0.925 & \revision{0.3} \\
\addlinespace[2pt]
CNXv2-L (FCMAE only) & 100 & 94.7\% & 54.9\% & 49.7\% & 94.0\% & 0.666 & \revision{1.3} \\
 & 200 & 94.7\% & 56.8\% & 53.0\% & 94.0\% & 0.752 & \revision{0.7} \\
 & 500 & 94.7\% & 56.9\% & 54.0\% & 94.0\% & 0.687 & \revision{0.7} \\
\addlinespace[2pt]
RN152 & 100 & 40.3\% & 69.1\% & 41.7\% & 61.0\% & 0.459 & \revision{58.3} \\
\bottomrule
\end{tabular}
\end{table*}

\clearpage

\begin{table*}[tp]
\centering
\caption{\revision{Complete synthetic results at $T{=}5000$ and $P{=}0.2$: all 12 backbones and all three concept-training budgets. Accuracies and label-switch rates are percentages; confidence is unitless.}}
\label{tab:app_sweep_T5000_P02}
\setlength{\tabcolsep}{2pt}
\footnotesize
\begin{tabular}{lrcccccc}
\toprule
\revision{\textbf{Backbone}} & \revision{$C$} & \revision{\textbf{LabelAcc}} & \revision{\textbf{ConceptAcc}} & \revision{\textbf{RuleAcc}} & \revision{\textbf{RuleLabelAcc}} & \revision{\textbf{Avg Conf}} & \revision{\textbf{Switch\%}} \\
\midrule
\revision{CNXv2-L} & \revision{100} & \revision{99.3\%} & \revision{90.0\%} & \revision{90.0\%} & \revision{99.3\%} & \revision{0.907} & \revision{0.0} \\
 & \revision{200} & \revision{99.3\%} & \revision{91.6\%} & \revision{92.3\%} & \revision{99.7\%} & \revision{0.902} & \revision{0.3} \\
 & \revision{500} & \revision{99.3\%} & \revision{93.3\%} & \revision{93.7\%} & \revision{99.7\%} & \revision{0.922} & \revision{0.3} \\
\addlinespace[2pt]
\revision{CNXv2-L (FCMAE only)} & \revision{100} & \revision{94.7\%} & \revision{54.9\%} & \revision{48.3\%} & \revision{94.0\%} & \revision{0.657} & \revision{1.3} \\
 & \revision{200} & \revision{94.7\%} & \revision{56.8\%} & \revision{53.0\%} & \revision{94.0\%} & \revision{0.754} & \revision{0.7} \\
 & \revision{500} & \revision{94.7\%} & \revision{56.9\%} & \revision{53.0\%} & \revision{94.0\%} & \revision{0.684} & \revision{0.7} \\
\addlinespace[2pt]
\revision{RN152} & \revision{100} & \revision{40.3\%} & \revision{69.1\%} & \revision{43.0\%} & \revision{63.0\%} & \revision{0.456} & \revision{57.7} \\
\bottomrule
\end{tabular}
\end{table*}

\clearpage

\subsection{\revision{Prior-estimation fraction}}
\label{app:psens}

\revision{Table~\ref{tab:app_psens} compares the two prior-estimation fractions while holding the label and concept predictors fixed. It gives RuleAcc at $T{=}5000,C{=}500$ and the maximum absolute change across all nine $(T,C)$ pairs. Tables~\ref{tab:app_sweep_T500}--\ref{tab:app_sweep_T5000_P02} report both fractions in full; no configuration is represented only by this sensitivity summary. The comparison measures sensitivity to the two estimated priors, rather than to arbitrary changes in concept prevalence.}

\revision{The synthetic sweep compares $P{=}0.1$ with $P{=}0.2$ using the same label and concept predictors. Their accuracies are unchanged. The maximum change in rule accuracy across the nine $(T,C)$ settings is at most $2.0$ percentage points for 10 of the 12 backbones ($0.3$ points for CNXv2-L). The exceptions are ViT-B/16~(SigLIP\,2) and RN50~(TV), with maxima of $5.7$ and $7.0$ points. Mean confidence changes by less than $0.02$ for every backbone. These comparisons support limited sensitivity for most evaluated configurations; they do not establish a general sample-complexity guarantee for prior estimation.}

\revision{Table~\ref{tab:app_psens} reports the rule accuracies at $T{=}5000,C{=}500$ and maximum absolute changes. At $T{=}5000,C{=}500$ the two samples' estimated concept priors differ by at most $0.012$. The bee results have identical aggregate metrics at the displayed precision for the two fractions; Appendix~\ref{app:real_results} describes their prior-fraction comparison.}

\begin{table*}[tp]
\centering
\caption{\revision{Prior-fraction sensitivity. The two RuleAcc columns use $T{=}5000,C{=}500$; maximum absolute accuracy changes across the nine $(T,C)$ pairs are in percentage points. LabelAcc and ConceptAcc are unchanged.}}
\label{tab:app_psens}
\setlength{\tabcolsep}{2pt}
\footnotesize
\begin{tabular}{lccccc}
\toprule
 & \multicolumn{2}{c}{\revision{RuleAcc, $T{=}5000,C{=}500$}} & \multicolumn{3}{c}{\revision{Maximum absolute change}} \\
Backbone & $P{=}0.1$ & $P{=}0.2$ & \revision{RuleAcc (pp)} & \revision{RuleLabelAcc (pp)} & \revision{Avg Conf} \\
\midrule
CNXv2-L & 93.7\% & 93.7\% & \revision{0.3} & \revision{0.0} & 0.003 \\
CNXv2-L (FCMAE only) & 54.0\% & 53.0\% & \revision{1.7} & \revision{0.0} & 0.009 \\
RN152 & 47.7\% & 47.3\% & \revision{1.3} & \revision{2.0} & 0.004 \\
RN50 (TV) & 69.7\% & 62.7\% & \revision{7.0} & \revision{7.7} & 0.005 \\
RN50 (A1h) & 67.7\% & 68.0\% & \revision{0.7} & \revision{1.3} & 0.007 \\
ViT-B/16 (DINO) & 96.7\% & 96.7\% & \revision{0.3} & \revision{0.0} & 0.003 \\
ViT-B/16 (MAE) & 75.0\% & 75.3\% & \revision{1.0} & \revision{0.3} & 0.007 \\
ViT-B/16 (SigLIP\,2) & 90.0\% & 90.0\% & \revision{5.7} & \revision{0.0} & 0.006 \\
BEiT-L/16 & 91.3\% & 91.3\% & \revision{1.0} & \revision{0.0} & 0.003 \\
Swin-B (IN-1k) & 90.7\% & 90.7\% & \revision{2.0} & \revision{0.3} & 0.003 \\
Swin-B (IN-22k) & 94.7\% & 93.7\% & \revision{1.0} & \revision{0.0} & 0.004 \\
SwinV2-S & 94.0\% & 93.7\% & \revision{0.3} & \revision{0.0} & 0.003 \\
\bottomrule
\end{tabular}
\end{table*}

\clearpage
\subsection{\revision{Per-concept synthetic test accuracies}}
\label{app:synthetic_concepts}

\revision{The following tables report the 16 concept accuracies for each of the 108 $(\text{backbone},T,C)$ combinations. The two prior-estimation fractions share these predictors, so each accuracy is reported once. Every concept is tested on 200 balanced examples. The arithmetic mean across both eight-concept panels for a row gives ConceptAcc in the benchmark tables.}

\begin{table*}[tp]
\centering
\caption{\revision{Synthetic concept-test accuracy (\%) at $T{=}500$, panel 1 of 2. $C$ is the number of positive and negative training examples per concept. Values are shared by $P{=}0.1$ and $P{=}0.2$.}}
\label{tab:synthetic_concepts_T500_1}
\setlength{\tabcolsep}{3pt}
\footnotesize
\begin{tabular}{lrcccccccc}
\toprule
\revision{Backbone} & \revision{$C$} & \revision{Alorn} & \revision{Floom} & \revision{Frib} & \revision{Guffit} & \revision{Hebda} & \revision{Heffix} & \revision{Krad} & \revision{Opple} \\
\midrule
\revision{CNXv2-L} & \revision{100} & \revision{86.0} & \revision{91.5} & \revision{76.0} & \revision{94.0} & \revision{98.5} & \revision{95.5} & \revision{89.5} & \revision{88.5} \\
 & \revision{200} & \revision{90.0} & \revision{93.0} & \revision{81.0} & \revision{92.0} & \revision{98.0} & \revision{94.0} & \revision{92.0} & \revision{90.5} \\
 & \revision{500} & \revision{93.5} & \revision{95.5} & \revision{85.0} & \revision{94.0} & \revision{99.0} & \revision{96.5} & \revision{92.5} & \revision{89.5} \\
\addlinespace[2pt]
\revision{CNXv2-L (FCMAE only)} & \revision{100} & \revision{58.5} & \revision{56.0} & \revision{50.0} & \revision{62.0} & \revision{61.5} & \revision{50.5} & \revision{58.0} & \revision{55.0} \\
 & \revision{200} & \revision{72.0} & \revision{54.0} & \revision{57.0} & \revision{72.0} & \revision{62.0} & \revision{51.5} & \revision{64.0} & \revision{52.0} \\
 & \revision{500} & \revision{61.5} & \revision{48.5} & \revision{51.5} & \revision{80.5} & \revision{65.5} & \revision{62.0} & \revision{52.0} & \revision{55.0} \\
\addlinespace[2pt]
\revision{RN152} & \revision{100} & \revision{70.0} & \revision{60.0} & \revision{70.0} & \revision{76.5} & \revision{66.5} & \revision{64.0} & \revision{65.5} & \revision{65.0} \\
\bottomrule
\end{tabular}
\end{table*}

\clearpage

\begin{table*}[tp]
\centering
\caption{\revision{Synthetic concept-test accuracy (\%) at $T{=}500$, panel 2 of 2. $C$ is the number of positive and negative training examples per concept. Values are shared by $P{=}0.1$ and $P{=}0.2$.}}
\label{tab:synthetic_concepts_T500_2}
\setlength{\tabcolsep}{3pt}
\footnotesize
\begin{tabular}{lrcccccccc}
\toprule
\revision{Backbone} & \revision{$C$} & \revision{Smeek} & \revision{Sorbax} & \revision{Voop} & \revision{Vornex} & \revision{Wexil} & \revision{Yintar} & \revision{Zandool} & \revision{Zorf} \\
\midrule
\revision{CNXv2-L} & \revision{100} & \revision{83.5} & \revision{83.0} & \revision{94.5} & \revision{97.0} & \revision{90.0} & \revision{92.0} & \revision{95.0} & \revision{82.5} \\
 & \revision{200} & \revision{88.0} & \revision{86.0} & \revision{95.0} & \revision{91.5} & \revision{94.0} & \revision{96.5} & \revision{96.5} & \revision{88.5} \\
 & \revision{500} & \revision{87.5} & \revision{88.0} & \revision{98.0} & \revision{97.5} & \revision{98.5} & \revision{98.0} & \revision{98.0} & \revision{92.0} \\
\addlinespace[2pt]
\revision{CNXv2-L (FCMAE only)} & \revision{100} & \revision{61.5} & \revision{57.5} & \revision{51.0} & \revision{52.0} & \revision{53.5} & \revision{52.5} & \revision{54.0} & \revision{49.0} \\
 & \revision{200} & \revision{64.5} & \revision{57.5} & \revision{48.5} & \revision{54.0} & \revision{64.0} & \revision{50.5} & \revision{62.0} & \revision{50.5} \\
 & \revision{500} & \revision{75.5} & \revision{55.0} & \revision{54.0} & \revision{51.5} & \revision{51.0} & \revision{53.0} & \revision{55.0} & \revision{53.5} \\
\addlinespace[2pt]
\revision{RN152} & \revision{100} & \revision{70.5} & \revision{70.5} & \revision{62.0} & \revision{69.5} & \revision{68.5} & \revision{63.5} & \revision{72.5} & \revision{59.0} \\
\bottomrule
\end{tabular}
\end{table*}

\clearpage

\begin{table*}[tp]
\centering
\caption{\revision{Synthetic concept-test accuracy (\%) at $T{=}2000$, panel 1 of 2. $C$ is the number of positive and negative training examples per concept. Values are shared by $P{=}0.1$ and $P{=}0.2$.}}
\label{tab:synthetic_concepts_T2000_1}
\setlength{\tabcolsep}{3pt}
\footnotesize
\begin{tabular}{lrcccccccc}
\toprule
\revision{Backbone} & \revision{$C$} & \revision{Alorn} & \revision{Floom} & \revision{Frib} & \revision{Guffit} & \revision{Hebda} & \revision{Heffix} & \revision{Krad} & \revision{Opple} \\
\midrule
\revision{CNXv2-L} & \revision{100} & \revision{85.0} & \revision{90.5} & \revision{79.5} & \revision{95.5} & \revision{98.5} & \revision{96.0} & \revision{89.5} & \revision{85.5} \\
 & \revision{200} & \revision{88.0} & \revision{85.0} & \revision{79.5} & \revision{92.0} & \revision{99.0} & \revision{96.5} & \revision{89.5} & \revision{89.5} \\
 & \revision{500} & \revision{90.5} & \revision{94.5} & \revision{85.0} & \revision{95.0} & \revision{98.5} & \revision{95.5} & \revision{95.5} & \revision{93.0} \\
\addlinespace[2pt]
\revision{CNXv2-L (FCMAE only)} & \revision{100} & \revision{58.5} & \revision{52.0} & \revision{52.0} & \revision{61.0} & \revision{59.5} & \revision{52.5} & \revision{53.5} & \revision{58.0} \\
 & \revision{200} & \revision{68.0} & \revision{52.0} & \revision{56.0} & \revision{72.5} & \revision{63.0} & \revision{54.0} & \revision{59.0} & \revision{52.0} \\
 & \revision{500} & \revision{58.5} & \revision{49.0} & \revision{53.0} & \revision{73.0} & \revision{63.5} & \revision{62.0} & \revision{52.0} & \revision{55.0} \\
\addlinespace[2pt]
\revision{RN152} & \revision{100} & \revision{74.0} & \revision{61.5} & \revision{71.5} & \revision{72.5} & \revision{64.5} & \revision{66.0} & \revision{67.5} & \revision{66.0} \\
\bottomrule
\end{tabular}
\end{table*}

\clearpage

\begin{table*}[tp]
\centering
\caption{\revision{Synthetic concept-test accuracy (\%) at $T{=}2000$, panel 2 of 2. $C$ is the number of positive and negative training examples per concept. Values are shared by $P{=}0.1$ and $P{=}0.2$.}}
\label{tab:synthetic_concepts_T2000_2}
\setlength{\tabcolsep}{3pt}
\footnotesize
\begin{tabular}{lrcccccccc}
\toprule
\revision{Backbone} & \revision{$C$} & \revision{Smeek} & \revision{Sorbax} & \revision{Voop} & \revision{Vornex} & \revision{Wexil} & \revision{Yintar} & \revision{Zandool} & \revision{Zorf} \\
\midrule
\revision{CNXv2-L} & \revision{100} & \revision{85.0} & \revision{83.5} & \revision{94.5} & \revision{97.5} & \revision{91.5} & \revision{92.5} & \revision{96.0} & \revision{84.0} \\
 & \revision{200} & \revision{87.5} & \revision{82.5} & \revision{94.5} & \revision{94.0} & \revision{94.5} & \revision{97.5} & \revision{95.5} & \revision{92.0} \\
 & \revision{500} & \revision{89.5} & \revision{90.0} & \revision{99.0} & \revision{95.5} & \revision{97.5} & \revision{96.5} & \revision{98.0} & \revision{92.5} \\
\addlinespace[2pt]
\revision{CNXv2-L (FCMAE only)} & \revision{100} & \revision{67.0} & \revision{57.0} & \revision{51.5} & \revision{51.0} & \revision{54.0} & \revision{51.0} & \revision{53.5} & \revision{49.0} \\
 & \revision{200} & \revision{59.0} & \revision{56.5} & \revision{48.5} & \revision{57.5} & \revision{57.0} & \revision{51.5} & \revision{52.5} & \revision{55.0} \\
 & \revision{500} & \revision{66.0} & \revision{58.0} & \revision{55.0} & \revision{50.5} & \revision{55.0} & \revision{52.0} & \revision{61.5} & \revision{53.5} \\
\addlinespace[2pt]
\revision{RN152} & \revision{100} & \revision{65.5} & \revision{61.5} & \revision{61.0} & \revision{80.5} & \revision{69.0} & \revision{55.5} & \revision{77.0} & \revision{58.0} \\
\bottomrule
\end{tabular}
\end{table*}

\clearpage

\begin{table*}[tp]
\centering
\caption{\revision{Synthetic concept-test accuracy (\%) at $T{=}5000$, panel 1 of 2. $C$ is the number of positive and negative training examples per concept. Values are shared by $P{=}0.1$ and $P{=}0.2$.}}
\label{tab:synthetic_concepts_T5000_1}
\setlength{\tabcolsep}{3pt}
\footnotesize
\begin{tabular}{lrcccccccc}
\toprule
\revision{Backbone} & \revision{$C$} & \revision{Alorn} & \revision{Floom} & \revision{Frib} & \revision{Guffit} & \revision{Hebda} & \revision{Heffix} & \revision{Krad} & \revision{Opple} \\
\midrule
\revision{CNXv2-L} & \revision{100} & \revision{84.5} & \revision{89.0} & \revision{81.5} & \revision{95.5} & \revision{95.5} & \revision{96.5} & \revision{90.0} & \revision{86.5} \\
 & \revision{200} & \revision{91.5} & \revision{92.0} & \revision{82.0} & \revision{92.5} & \revision{99.0} & \revision{93.0} & \revision{91.0} & \revision{93.0} \\
 & \revision{500} & \revision{93.5} & \revision{89.0} & \revision{83.0} & \revision{95.5} & \revision{98.5} & \revision{94.5} & \revision{95.5} & \revision{88.0} \\
\addlinespace[2pt]
\revision{CNXv2-L (FCMAE only)} & \revision{100} & \revision{57.5} & \revision{54.0} & \revision{51.0} & \revision{63.0} & \revision{57.0} & \revision{50.0} & \revision{52.0} & \revision{59.0} \\
 & \revision{200} & \revision{64.5} & \revision{58.0} & \revision{55.5} & \revision{72.5} & \revision{61.0} & \revision{50.5} & \revision{63.5} & \revision{51.5} \\
 & \revision{500} & \revision{62.5} & \revision{50.5} & \revision{53.5} & \revision{77.5} & \revision{61.0} & \revision{62.0} & \revision{48.0} & \revision{56.0} \\
\addlinespace[2pt]
\revision{RN152} & \revision{100} & \revision{75.0} & \revision{68.0} & \revision{73.0} & \revision{74.0} & \revision{65.0} & \revision{67.5} & \revision{59.5} & \revision{65.0} \\
\bottomrule
\end{tabular}
\end{table*}

\clearpage

\begin{table*}[tp]
\centering
\caption{\revision{Synthetic concept-test accuracy (\%) at $T{=}5000$, panel 2 of 2. $C$ is the number of positive and negative training examples per concept. Values are shared by $P{=}0.1$ and $P{=}0.2$.}}
\label{tab:synthetic_concepts_T5000_2}
\setlength{\tabcolsep}{3pt}
\footnotesize
\begin{tabular}{lrcccccccc}
\toprule
\revision{Backbone} & \revision{$C$} & \revision{Smeek} & \revision{Sorbax} & \revision{Voop} & \revision{Vornex} & \revision{Wexil} & \revision{Yintar} & \revision{Zandool} & \revision{Zorf} \\
\midrule
\revision{CNXv2-L} & \revision{100} & \revision{84.0} & \revision{85.0} & \revision{94.5} & \revision{96.5} & \revision{93.0} & \revision{91.0} & \revision{93.5} & \revision{83.5} \\
 & \revision{200} & \revision{87.0} & \revision{85.0} & \revision{93.5} & \revision{91.5} & \revision{91.0} & \revision{97.0} & \revision{94.5} & \revision{91.5} \\
 & \revision{500} & \revision{89.5} & \revision{85.0} & \revision{97.5} & \revision{96.5} & \revision{98.0} & \revision{97.0} & \revision{97.5} & \revision{94.5} \\
\addlinespace[2pt]
\revision{CNXv2-L (FCMAE only)} & \revision{100} & \revision{63.0} & \revision{57.0} & \revision{54.0} & \revision{50.5} & \revision{54.5} & \revision{51.5} & \revision{53.5} & \revision{50.5} \\
 & \revision{200} & \revision{58.5} & \revision{56.0} & \revision{49.0} & \revision{56.0} & \revision{60.0} & \revision{51.0} & \revision{53.0} & \revision{48.0} \\
 & \revision{500} & \revision{63.5} & \revision{56.0} & \revision{56.0} & \revision{52.0} & \revision{51.0} & \revision{53.0} & \revision{56.0} & \revision{52.0} \\
\addlinespace[2pt]
\revision{RN152} & \revision{100} & \revision{71.5} & \revision{65.0} & \revision{65.5} & \revision{80.5} & \revision{73.5} & \revision{57.0} & \revision{79.0} & \revision{67.0} \\
\bottomrule
\end{tabular}
\end{table*}

\clearpage

\subsection{\revision{Synthetic label-head accuracies}}
\label{app:synthetic_heads}

\revision{Table~\ref{tab:synthetic_heads} reports the validation and test accuracies of every synthetic label head. These splits are distinct from the 300-image Explanation Test split used for LabelAcc in the benchmark tables. Validation accuracy is available as a fraction rounded to two decimals in the classification reports and is therefore shown to the nearest percentage point; test accuracy is computed from exact confusion counts. At fixed $T$, the same label head is used for all three concept-training budgets and both prior-estimation fractions.}

\begin{table*}[tp]
\centering
\caption{\revision{Synthetic label-head accuracy (\%) on the validation and test splits. The test split contains 3\,000 images; validation has 75, 300, and 750 images at $T{=}500,2000,5000$, respectively. All 36 trained heads are shown.}}
\label{tab:synthetic_heads}
\setlength{\tabcolsep}{3pt}
\footnotesize
\begin{tabular}{lcccccc}
\toprule
 & \multicolumn{2}{c}{\revision{$T{=}500$}} & \multicolumn{2}{c}{\revision{$T{=}2000$}} & \multicolumn{2}{c}{\revision{$T{=}5000$}} \\
\revision{Backbone} & \revision{Validation} & \revision{Test} & \revision{Validation} & \revision{Test} & \revision{Validation} & \revision{Test} \\
\midrule
\revision{CNXv2-L} & \revision{100\%} & \revision{99.0\%} & \revision{100\%} & \revision{99.4\%} & \revision{100\%} & \revision{99.5\%} \\
\revision{CNXv2-L (FCMAE only)} & \revision{89\%} & \revision{93.0\%} & \revision{95\%} & \revision{95.4\%} & \revision{97\%} & \revision{96.6\%} \\
\revision{RN152} & \revision{49\%} & \revision{43.9\%} & \revision{42\%} & \revision{43.5\%} & \revision{40\%} & \revision{41.5\%} \\
\revision{RN50 (TV)} & \revision{43\%} & \revision{37.7\%} & \revision{33\%} & \revision{34.0\%} & \revision{39\%} & \revision{37.0\%} \\
\revision{RN50 (A1h)} & \revision{60\%} & \revision{49.3\%} & \revision{52\%} & \revision{49.6\%} & \revision{53\%} & \revision{49.0\%} \\
\revision{ViT-B/16 (DINO)} & \revision{100\%} & \revision{99.6\%} & \revision{99\%} & \revision{99.2\%} & \revision{100\%} & \revision{99.8\%} \\
\revision{ViT-B/16 (MAE)} & \revision{96\%} & \revision{93.9\%} & \revision{99\%} & \revision{98.2\%} & \revision{99\%} & \revision{98.9\%} \\
\revision{ViT-B/16 (SigLIP\,2)} & \revision{99\%} & \revision{99.3\%} & \revision{99\%} & \revision{99.5\%} & \revision{100\%} & \revision{99.7\%} \\
\revision{BEiT-L/16} & \revision{99\%} & \revision{99.0\%} & \revision{99\%} & \revision{99.1\%} & \revision{100\%} & \revision{99.6\%} \\
\revision{Swin-B (IN-1k)} & \revision{100\%} & \revision{98.6\%} & \revision{99\%} & \revision{99.1\%} & \revision{99\%} & \revision{99.5\%} \\
\revision{Swin-B (IN-22k)} & \revision{100\%} & \revision{99.0\%} & \revision{100\%} & \revision{99.1\%} & \revision{100\%} & \revision{99.6\%} \\
\revision{SwinV2-S} & \revision{100\%} & \revision{97.5\%} & \revision{99\%} & \revision{98.2\%} & \revision{99\%} & \revision{99.5\%} \\
\bottomrule
\end{tabular}
\end{table*}

\clearpage
\section{Complete Bee Results and Predictor Accuracies}
\label{app:real_results}

\revision{The \emph{bee reference configuration} uses the fixed dataset splits described in Appendix~\ref{sec:real_dataset}, prior-estimation fraction $P{=}0.1$, and seed~42. This appendix reports all 24 backbone/prior-fraction combinations: 12 backbones at each of $P\in\{0.1,0.2\}$. It also reports the individual concept and label-head accuracies. All results are single-run measurements. Accuracies are rounded half-up to one decimal and confidence to three decimals; full-precision reports and counts are supplied separately in \texttt{experiment\_results}.}

\revision{LabelAcc, RuleAcc, RuleLabelAcc, and Avg~$\operatorname{Conf}$ use the 600-image Explanation Test split. ConceptAcc is the unweighted mean of the four concept-test accuracies. There is one rule per class, so RuleAcc and RuleLabelAcc coincide. The overlap and class-template limitations in Appendix~\ref{sec:real_dataset} apply: these results do not constitute an independent evaluation on new observations or independently annotated visible concepts.}

\subsection{\revision{Complete bee benchmark results}}

\begin{table*}[tp]
\centering
\caption{\revision{All 12 backbones in the bee reference configuration ($P{=}0.1$, seed~42). Accuracies are percentages; confidence is unitless. Bold marks column maxima. The split-overlap limitations in Appendix~\ref{sec:real_dataset} apply.}}
\label{tab:real_auto}
\setlength{\tabcolsep}{3pt}
\footnotesize
\begin{tabular}{lccccc}
\toprule
\textbf{Backbone} & \revision{\textbf{LabelAcc}} & \revision{\textbf{ConceptAcc}} & \revision{\textbf{RuleAcc}} & \revision{\textbf{RuleLabelAcc}} & \revision{\textbf{Avg Conf}} \\
\midrule
CNXv2-L & \textbf{99.7\%} & 96.4\% & 85.8\% & 85.8\% & \textbf{0.967} \\
CNXv2-L (FCMAE only) & 63.3\% & 68.3\% & 43.8\% & 43.8\% & 0.633 \\
RN152 & 84.7\% & 87.0\% & 62.8\% & 62.8\% & 0.803 \\
RN50 (TV) & 91.2\% & 85.5\% & 87.8\% & 87.8\% & 0.770 \\
RN50 (A1h) & 87.0\% & 87.0\% & 66.7\% & 66.7\% & 0.792 \\
ViT-B/16 (DINO) & 97.7\% & 96.2\% & 84.3\% & 84.3\% & 0.927 \\
ViT-B/16 (MAE) & 81.5\% & 85.5\% & 66.2\% & 66.2\% & 0.793 \\
ViT-B/16 (SigLIP\,2) & 78.2\% & 84.4\% & 58.5\% & 58.5\% & 0.763 \\
BEiT-L/16 & 80.8\% & 84.1\% & 63.2\% & 63.2\% & 0.798 \\
Swin-B (IN-1k) & 96.0\% & 95.3\% & 82.2\% & 82.2\% & 0.946 \\
Swin-B (IN-22k) & \textbf{99.7\%} & \textbf{97.7\%} & \textbf{98.2\%} & \textbf{98.2\%} & 0.947 \\
SwinV2-S & 97.3\% & 95.2\% & 83.5\% & 83.5\% & 0.943 \\
\bottomrule
\end{tabular}
\end{table*}

\begin{table*}[tp]
\centering
\caption{\revision{Complete bee results at $P{=}0.2$, seed~42. All five metrics are shown for every backbone. At the displayed precision they equal the corresponding $P{=}0.1$ values.}}
\label{tab:real_auto_P02}
\setlength{\tabcolsep}{3pt}
\footnotesize
\begin{tabular}{lccccc}
\toprule
\revision{\textbf{Backbone}} & \revision{\textbf{LabelAcc}} & \revision{\textbf{ConceptAcc}} & \revision{\textbf{RuleAcc}} & \revision{\textbf{RuleLabelAcc}} & \revision{\textbf{Avg Conf}} \\
\midrule
\revision{CNXv2-L} & \revision{\textbf{99.7\%}} & \revision{96.4\%} & \revision{85.8\%} & \revision{85.8\%} & \revision{\textbf{0.967}} \\
\revision{CNXv2-L (FCMAE only)} & \revision{63.3\%} & \revision{68.3\%} & \revision{43.8\%} & \revision{43.8\%} & \revision{0.633} \\
\revision{RN152} & \revision{84.7\%} & \revision{87.0\%} & \revision{62.8\%} & \revision{62.8\%} & \revision{0.803} \\
\revision{RN50 (TV)} & \revision{91.2\%} & \revision{85.5\%} & \revision{87.8\%} & \revision{87.8\%} & \revision{0.770} \\
\revision{RN50 (A1h)} & \revision{87.0\%} & \revision{87.0\%} & \revision{66.7\%} & \revision{66.7\%} & \revision{0.792} \\
\revision{ViT-B/16 (DINO)} & \revision{97.7\%} & \revision{96.2\%} & \revision{84.3\%} & \revision{84.3\%} & \revision{0.927} \\
\revision{ViT-B/16 (MAE)} & \revision{81.5\%} & \revision{85.5\%} & \revision{66.2\%} & \revision{66.2\%} & \revision{0.793} \\
\revision{ViT-B/16 (SigLIP\,2)} & \revision{78.2\%} & \revision{84.4\%} & \revision{58.5\%} & \revision{58.5\%} & \revision{0.763} \\
\revision{BEiT-L/16} & \revision{80.8\%} & \revision{84.1\%} & \revision{63.2\%} & \revision{63.2\%} & \revision{0.798} \\
\revision{Swin-B (IN-1k)} & \revision{96.0\%} & \revision{95.3\%} & \revision{82.2\%} & \revision{82.2\%} & \revision{0.946} \\
\revision{Swin-B (IN-22k)} & \revision{\textbf{99.7\%}} & \revision{\textbf{97.7\%}} & \revision{\textbf{98.2\%}} & \revision{\textbf{98.2\%}} & \revision{0.947} \\
\revision{SwinV2-S} & \revision{97.3\%} & \revision{95.2\%} & \revision{83.5\%} & \revision{83.5\%} & \revision{0.943} \\
\bottomrule
\end{tabular}
\end{table*}

\clearpage
\subsection{Detailed bee findings}

\revision{{\emph{Agreement with the bee rules varies across backbones.} On the 600-image Explanation Test partition, rule accuracy ranges from $43.8\%$ (CNXv2-L, FCMAE only; 263 correct) to $98.2\%$ (Swin-B, IN-22k; 589 correct). The five backbones with at least $96\%$ label accuracy on this partition (Swin-B~(IN-22k), CNXv2-L, ViT-B/16~(DINO), SwinV2-S, Swin-B~(IN-1k)) agree with the class-associated rule for $82.2$--$98.2\%$ of images, with mean normalized rule confidence $0.927$--$0.967$. CNXv2-L has $99.7\%$ label accuracy, $85.8\%$ rule accuracy, and mean confidence $0.967$. For these five backbones, individual concept-classifier accuracies are $90.2$--$99.3\%$ (means $95.2$--$97.7\%$; CNXv2-L $93.2$--$98.6\%$, mean $96.4\%$). The concept sets contain same-source training/test overlap, and the benchmark vectors are class templates, so these measurements do not establish independent recognition of each cue on the benchmark photographs.}}

{\emph{Rule selection can reduce label accuracy on field photographs.} Because each class has exactly one rule, every rule error is also a rule-label error, and the gap between label accuracy and rule accuracy measures the net effect of rule selection relative to the classifier; this gap is present for every backbone, from $1.5$ points (Swin-B~(IN-22k), $99.7\%$ vs.\ $98.2\%$) to $21.8$ points (RN152, $84.7\%$ vs.\ $62.8\%$), and rule accuracy never exceeds label accuracy on this dataset. For CNXv2-L ($99.7\%$ vs.\ $85.8\%$), on $13.8\%$ of images (83 of 600) the scoring procedure in Equation~(\ref{eq_rule_post_restricted}) selects a rule of a different class from the (almost always correct) backbone prediction. Concept accuracy is measured on the curated concept exemplars, whereas Explanation Test images are unconstrained field photographs in which cues can be occluded, viewpoint-dependent, or transient (\eg visible pollen loads vary across photographs); concept accuracy on these images may therefore differ from the curated estimate, and errors can compound across the two to four literals per rule. The high mean confidence (Avg~${\operatorname{Conf}} = 0.967$) despite imperfect rule accuracy shows that decisive rule scores can accompany errors; these results do not validate a confidence threshold for presenting explanations.}

\revision{{\emph{Backbone quality and rule confidence vary on real data.} Some backbones have lower accuracy in both evidence channels. For RN152, label accuracy on the Explanation Test partition is $84.7\%$ (its fine-tuned head reaches $73.1\%$ on the Label Test split) and mean concept accuracy is $87.0\%$ ($81.4$--$92.9\%$ per concept), yielding a rule accuracy of $62.8\%$; CNXv2-L with FCMAE pretraining only is the weakest backbone on both channels ($63.3\%$ label, $68.3\%$ concept) and reaches $43.8\%$ rule accuracy. Rule-implied label accuracy is lower than classifier accuracy for every backbone on this dataset. Because there is one rule per class, rule accuracy and rule-implied label accuracy coincide; individual concept errors need not change the selected rule. Across backbones, rule accuracy follows the label channel more closely (Spearman $\rho = 0.90$ with label accuracy vs.\ $0.76$ with mean concept accuracy across the 12 backbones), the reverse of the synthetic ordering. The aggregate confidence pattern is: the five backbones with Avg~${\operatorname{Conf}} \geq 0.92$ all exceed $82\%$ rule accuracy, and six of the seven below $0.81$ fall below $67\%$; the exception is RN50~(TV), whose mean rule confidence is comparatively low ($0.770$) despite $87.8\%$ rule accuracy. The rank correlation between confidence and rule accuracy across backbones is $\rho = 0.64$ (Pearson $r = 0.84$), weaker than on the synthetic data. This association does not establish reliable confidence-based ranking for new deployments. The checkpoint comparison gives Swin-B $82.2\%$ and $98.2\%$ rule accuracy for IN-1k and IN-22k checkpoints, respectively; ViT-B/16 reaches $84.3\%$ with DINO against $66.2\%$ (MAE) and $58.5\%$ (SigLIP\,2), and the FCMAE-only and supervised-fine-tuned CNXv2-L checkpoints obtain $43.8\%$ and $85.8\%$, respectively. These comparisons use a single run per configuration, with no repeated-seed uncertainty estimates.}}

\begin{table*}[tp]
\centering
\caption{\revision{Bee benchmark counts out of 600 images. Every count is identical at $P{=}0.1$ and $P{=}0.2$. A label switch means that the selected rule implies a class different from the classifier prediction; it is not necessarily an additional error.}}
\label{tab:bee_counts}
\setlength{\tabcolsep}{3pt}
\footnotesize
\begin{tabular}{lrrrrr}
\toprule
\revision{Backbone} & \revision{Correct labels} & \revision{Correct rules} & \revision{Correct rule labels} & \revision{Switches} & \revision{Switch\%} \\
\midrule
\revision{CNXv2-L} & \revision{598} & \revision{515} & \revision{515} & \revision{83} & \revision{13.8} \\
\revision{CNXv2-L (FCMAE only)} & \revision{380} & \revision{263} & \revision{263} & \revision{219} & \revision{36.5} \\
\revision{RN152} & \revision{508} & \revision{377} & \revision{377} & \revision{163} & \revision{27.2} \\
\revision{RN50 (TV)} & \revision{547} & \revision{527} & \revision{527} & \revision{43} & \revision{7.2} \\
\revision{RN50 (A1h)} & \revision{522} & \revision{400} & \revision{400} & \revision{151} & \revision{25.2} \\
\revision{ViT-B/16 (DINO)} & \revision{586} & \revision{506} & \revision{506} & \revision{80} & \revision{13.3} \\
\revision{ViT-B/16 (MAE)} & \revision{489} & \revision{397} & \revision{397} & \revision{134} & \revision{22.3} \\
\revision{ViT-B/16 (SigLIP\,2)} & \revision{469} & \revision{351} & \revision{351} & \revision{213} & \revision{35.5} \\
\revision{BEiT-L/16} & \revision{485} & \revision{379} & \revision{379} & \revision{162} & \revision{27.0} \\
\revision{Swin-B (IN-1k)} & \revision{576} & \revision{493} & \revision{493} & \revision{91} & \revision{15.2} \\
\revision{Swin-B (IN-22k)} & \revision{598} & \revision{589} & \revision{589} & \revision{9} & \revision{1.5} \\
\revision{SwinV2-S} & \revision{584} & \revision{501} & \revision{501} & \revision{87} & \revision{14.5} \\
\bottomrule
\end{tabular}
\end{table*}

\clearpage
\subsection{\revision{Bee per-concept and label-head accuracies}}
\label{app:real_concepts}

\revision{Table~\ref{tab:app_real_concepts} gives every concept-test accuracy and the corresponding macro-average. The four concept tests contain 134 fuzzy-body, 184 dense-hind-leg, 118 pollen-on-hind-leg, and 141 green-head/thorax entries, respectively (577 entries total). The same concept predictors are used at both prior-estimation fractions. Label-head validation and test accuracies use 3\,643 and 8\,125 images, respectively, and therefore differ from LabelAcc on the 600-image Explanation Test split. Validation accuracy is recorded as a fraction rounded to two decimals and is shown to the nearest percentage point; test accuracy is computed from exact confusion counts. The Label Test split supplies the label-confusion likelihood and includes the Explanation Test images, as detailed in Appendix~\ref{sec:real_dataset}.}

\begin{table*}[tp]
\centering
\caption{\revision{Bee concept and label-head accuracies (\%). The four concept columns correspond to fuzzy body, dense hair on hind legs, pollen on hind legs, and green head/thorax. Mean is the unweighted mean over concepts. Validation and test refer to the label-head splits, containing 3\,643 and 8\,125 images, respectively. Validation accuracy is available only to the nearest percentage point.}}
\label{tab:app_real_concepts}
\setlength{\tabcolsep}{2pt}
\footnotesize
\begin{tabular}{lccccccc}
\toprule
 & \multicolumn{5}{c}{\revision{Concept accuracy}} & \multicolumn{2}{c}{\revision{Label-head accuracy}} \\
Backbone & Fuzzy & Dense & Pollen & \revision{Green} & Mean & \revision{Validation} & \revision{Test} \\
\midrule
CNXv2-L & \revision{97.8} & \revision{96.2} & \revision{93.2} & \revision{98.6} & \revision{96.4} & \revision{95} & \revision{95.4} \\
CNXv2-L (FCMAE only) & \revision{80.6} & \revision{73.4} & \revision{69.5} & \revision{49.6} & \revision{68.3} & \revision{59} & \revision{60.1} \\
RN152 & \revision{88.8} & \revision{84.8} & \revision{81.4} & \revision{92.9} & \revision{87.0} & \revision{75} & \revision{73.1} \\
RN50 (TV) & \revision{87.3} & \revision{78.3} & \revision{84.7} & \revision{91.5} & \revision{85.5} & \revision{80} & \revision{79.5} \\
RN50 (A1h) & \revision{93.3} & \revision{80.4} & \revision{81.4} & \revision{92.9} & \revision{87.0} & \revision{79} & \revision{77.7} \\
ViT-B/16 (DINO) & \revision{97.0} & \revision{95.7} & \revision{95.8} & \revision{96.5} & \revision{96.2} & \revision{91} & \revision{90.3} \\
ViT-B/16 (MAE) & \revision{94.0} & \revision{72.8} & \revision{86.4} & \revision{88.7} & \revision{85.5} & \revision{73} & \revision{72.9} \\
ViT-B/16 (SigLIP\,2) & \revision{91.0} & \revision{64.7} & \revision{88.1} & \revision{93.6} & \revision{84.4} & \revision{72} & \revision{71.7} \\
BEiT-L/16 & \revision{90.3} & \revision{75.5} & \revision{79.7} & \revision{90.8} & \revision{84.1} & \revision{74} & \revision{73.5} \\
Swin-B (IN-1k) & \revision{98.5} & \revision{91.8} & \revision{92.4} & \revision{98.6} & \revision{95.3} & \revision{90} & \revision{91.1} \\
Swin-B (IN-22k) & \revision{99.3} & \revision{95.7} & \revision{97.5} & \revision{98.6} & \revision{97.7} & \revision{95} & \revision{94.7} \\
SwinV2-S & \revision{96.3} & \revision{90.2} & \revision{95.8} & \revision{98.6} & \revision{95.2} & \revision{91} & \revision{90.4} \\
\bottomrule
\end{tabular}
\end{table*}

\subsection{Prior-fraction sensitivity}

For the bee experiment, the two prior samples contain 7\,290 and 14\,577 Label Train images, respectively. Their class-associated templates yield nearly identical marginal priors: the dense-hair, pollen, and green-head/thorax prior changes from $0.3334705075$ to $0.3334705358$, and fuzzy-body prevalence is the complement. All correctness and switch counts are unchanged, as are all metrics at the displayed precision. The largest absolute change in mean confidence is approximately $2.05\times10^{-8}$. An exhaustive comparison of the $3\times2^4=48$ possible predicted-label/concept-vector inputs for each backbone also finds no selected-rule change under the two priors and identical likelihood tables. This establishes insensitivity to these two almost identical prior estimates; it does not establish robustness to substantial prevalence shift.

Per-image confidence calibration and selective prediction are not evaluated. The aggregate results do not determine the accuracy that would follow from removing the overlaps; quantifying that effect requires experiments with independently partitioned data.

\revision{Tables~\ref{tab:real_auto} and~\ref{tab:real_auto_P02} report all benchmark metrics for the two fractions, and Table~\ref{tab:bee_counts} gives their correctness and switch counts.}

\end{document}